%% file: main.tex
\documentclass[10pt,twocolumn,letterpaper]{article}

\usepackage{multirow}
\usepackage{color, colortbl} 
\usepackage{pifont}
\usepackage{caption}
\usepackage{tabularray}
\usepackage{tabularx}
\usepackage{algorithm}
\usepackage{algorithmic}
\usepackage{soul}
\usepackage{bm}

\newcommand{\sota}{state-of-the-art\xspace}

\definecolor{highlight}{rgb}{0.95,0.95,1}

\usepackage[pagenumbers]{iccv} 


\definecolor{iccvblue}{rgb}{0.21,0.49,0.74}
\usepackage[pagebackref,breaklinks,colorlinks,allcolors=iccvblue]{hyperref}

\def\paperID{14528} 
\def\confName{ICCV}
\def\confYear{2025}

\title{M2N2V2: Multi-Modal Unsupervised and Training-free Interactive Segmentation}

\author{Markus Karmann$^1$ \ \ Peng-Tao Jiang$^2$  \ \  Bo Li$^2$ \ \ Onay Urfalioglu$^1$\\
$^1$\tt{Vivo Tech Research GmbH}\\ $^2$\tt{vivo Mobile Communication Co., Ltd, Shanghai, China.}
}
\begin{document}

\makeatletter
\g@addto@macro\@maketitle{
  \begin{figure}[H]
  \setlength{\linewidth}{\textwidth}
  \setlength{\hsize}{\textwidth}
  \centering
  \includegraphics[width=\textwidth]{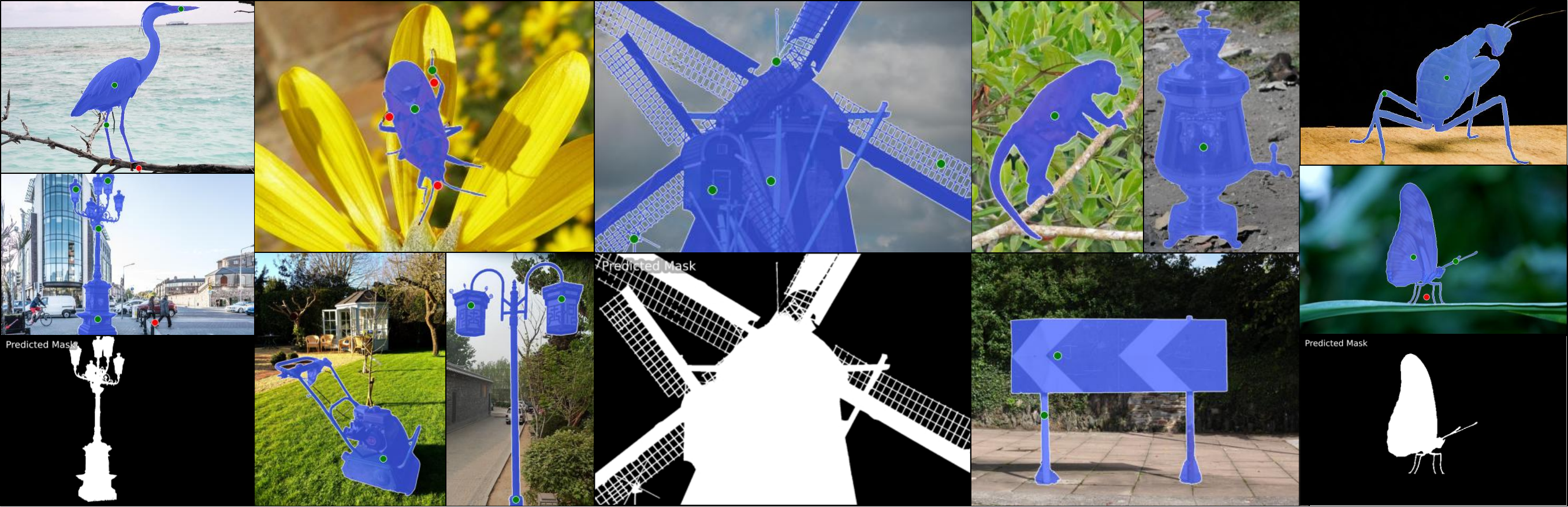}
  \caption{\textbf{We introduce M2N2V2, a multi-modal training-free point-prompt-based segmentation framework.} This figure shows qualitative segmentation results achieving an IoU $\ge 95\%$. Foreground prompt points are shown in green and background points in red.}
  \end{figure}
}
\makeatother

\maketitle
\input{sec/0_abstract}
\input{sec/1_introduction}
\input{sec/2_related_work}

\input{sec/3_method}

\input{sec/4_experiments}
\input{sec/5_conclusion}
{
    \small
    \bibliographystyle{ieeenat_fullname}
    \bibliography{main}
}

\input{sec/6_supplementary}

\end{document}

%% file: sec/0_abstract.tex
\begin{abstract}
We present {\bf M}arkov {\bf M}ap {\bf N}earest {\bf N}eighbor \textbf{V2} (M2N2V2), a novel and simple, yet effective approach which leverages depth guidance and attention maps for unsupervised and training-free point-prompt-based interactive segmentation.
Following recent trends in supervised multimodal approaches, we carefully integrate depth as an additional modality to create novel depth-guided Markov-maps.
Furthermore, we observe occasional segment size fluctuations in M2N2 during the interactive process, which can decrease the overall mIoU's.
To mitigate this problem, we model the prompting as a sequential process and propose a novel adaptive score function which considers the previous segmentation and the current prompt point in order to prevent unreasonable segment size changes.
Using Stable Diffusion 2 and Depth Anything V2 as backbones, we empirically show that our proposed M2N2V2 significantly improves the Number of Clicks (NoC) and mIoU compared to M2N2 in all datasets except those from the medical domain.
Interestingly, our unsupervised approach achieves competitive results compared to supervised methods like SAM and SimpleClick in the more challenging DAVIS and HQSeg44K datasets in the NoC metric, reducing the gap between supervised and unsupervised methods.
\end{abstract}

%% file: sec/1_introduction.tex
\section{Introduction}\label{sec:introduction}
\;\\
In point-prompt-based interactive segmentation, the user is shown an image together with an initially empty segmentation mask.
The user can then click on regions they either want to include (foreground point) or exclude (background point) in the desired segmentation.
Once the user places a new prompt point, the segmentation is updated based on all the points provided by the user so far.
This process is repeated until a satisfactory segmentation is achieved.
Recent advancements~\cite{M2N2} in unsupervised point-prompt-based segmentation show that by utilizing attention maps of Stable Diffusion 2 (SD2), it is possible to achieve promising results without training on manually or pseudo-labeled segmentation datasets.
Building upon M2N2~\cite{M2N2}, we propose M2N2V2 as a training-free approach which does not require any dataset.
Therefore, the segmentation performance does not depend on the quality of manually annotated segmentation labels.
We aim to reduce the gap between supervised training-based models and unsupervised training-free approaches by providing a framework that allows easy drop-in utilization of various backbones and future models.

Although attention maps provide very useful semantic information, their low resolution still poses a challenge in M2N2.
In order to achieve IoUs $\ge 95\%$, fine per-pixel details must be correctly segmented.
M2N2 relies on Joint Bilateral Upsampling (JBU)~\cite{JBU} with RGB as a guide to upscale the low resolution attention to image resolution.
This method improves segment boundaries and results in clearer edges, but only works if semantically similar regions also have similar colors and regions of different semantic classes have contrasting colors.
To mitigate this problem and inspired by OIS~\cite{OISegmentation}, we introduce depth as an additional modality and propose depth-guided Markov-maps.
We achieve this by integrating the depth information into the JBU as well as the flood fill of the Markov-map generation process of M2N2. 
In contrast to OIS, we directly utilize depth maps without transforming them first into order maps and show that a simple but careful integration without additional training can lead to significant improvements in the NoC and mIoU.

Inaccuracies in the Markov-maps can suddenly lead to undesired inclusion or exclusion of large regions in the segmentation when setting a new prompt point.
These cases often occur after setting a certain number of prompt points, where only small changes are required.
Typically, this involves placing prompt points close to the segment's boundary.
In this case, Markov-maps can fail to correctly distinguish between semantically adjacent regions, which can cause large erroneous changes in the segmentation and may require a significant amount of corrective prompts to be compensated.
This is often reflected in a sudden drop in mIoU.
As a countermeasure, we introduce a novel adaptive scoring function by measuring the distance of a new prompt point to the boundary of the previous segmentation.
This enables us to estimate the user-intended change of the segment area.
Using it as an additional control measure, we are able to prevent sudden and undesired collapses or expansions of the predicted segmentation. In general, our main contributions are the following.
\begin{itemize}
    \item We introduce \textbf{M2N2V2}, a novel training-free unsupervised framework for point-prompt-based interactive segmentation.
    \item We propose a novel method for creating \textbf{depth-guided Markov-maps}.
    \item We stabilize the interactive segmentation process by replacing the constant prior score function of M2N2 with a novel \textbf{adaptive segment size score function}.
    \item We perform extensive experiments on \textbf{10} datasets, achieving state-of-the-art results and surpassing M2N2 in all non-medical datasets and provide detailed analysis on strengths and limitations of our approach.
    \item We provide a full code release of our framework implemented in Python and PyTorch.
\end{itemize}

%% file: sec/2_related_work.tex
\section{Related Work}\label{sec:related_work}

\subsection{Supervised Methods}

An important branch of popular supervised methods was created by~\cite{DIOS}, where point prompts are modeled as distance maps and various point sampling strategies are used in a FCN~\cite{FCNSemanticSegmentation} framework.
More recent approaches include ~\cite{ritm2022}, ~\cite{interactiveSegmentationFirstClickAttn},~\cite{SAM, HQSAM}, ~\cite{Rana:Dynamite}, ~\cite{interactiveSegmentationbackpropRefinement, interactiveSegmentationfBRS}, ~\cite{SimpleClick}, 
~\cite{interactiveSegmentationGaussianProcess}, ~\cite{interactiveSegmentationFocusCut, FocalClick}, 
~\cite{clickPromptOptimalTransport}, ~\cite{CFR-ICL}, ~\cite{SegNext-InteractiveSegmentation},~\cite{interformer}, either 
deploying dense fusion, where prompts are encoded as prompt maps, or sparse fusion, where prompts are transformed into embedding space to be fused with image and other related embeddings.
Recently, OIS~\cite{OISegmentation} proposed order maps obtained from depth maps to enhance the prompt’s ability to separate segments more effectively, resulting in \sota NoC's and demonstrating how one can benefit from multi-modality. 

Generally, supervised methods require datasets with semantic labels, which are typically difficult to obtain in large amount and good quality. Furthermore, typically, unsupervised methods tend to generalize better than supervised methods.  

\begin{figure*}[!ht]
  \centering
  \includegraphics[width=\textwidth]{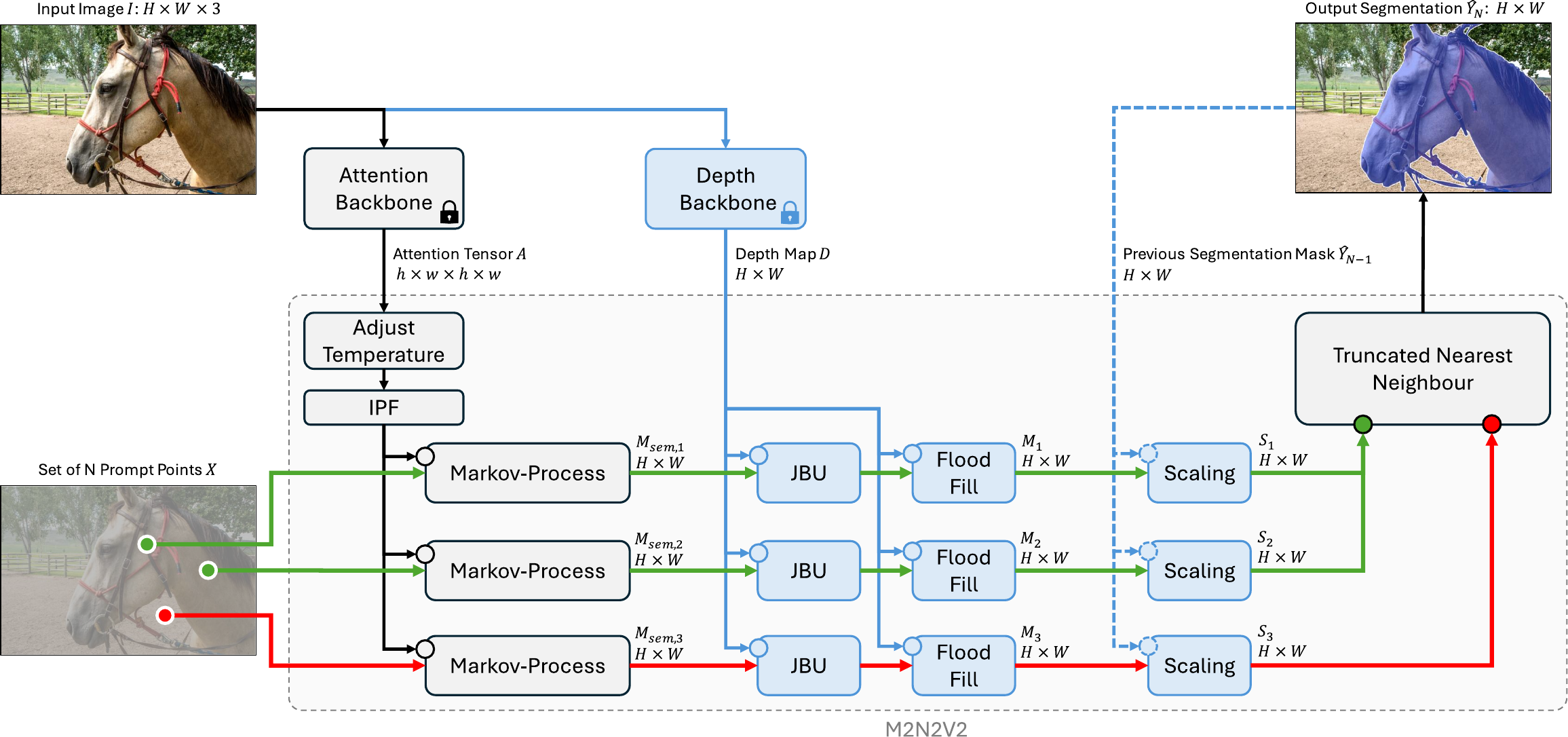}
  \caption{\textbf{M2N2V2 framework overview}. Our contributions are highlighted in blue. The illustration shows an example with $N=3$ prompt points (2 green foreground points, 1 red background point). First, we extract the attention tensor $\boldsymbol{A}$ and depth map $D$ of a given image $I$. Then we use the depth, RGB and attention information to generate Markov-maps $M_i$ for each prompt point $x_i$. Finally, each Markov-map is scaled based on a scoring function utilizing the previous segmentation result $\hat{Y}_{N-1}$ to predict $\hat{Y}_N$.}
  \label{fig:framework_overview}
\end{figure*}

\subsection{Unsupervised methods}
Non-deep-learning-based unsupervised methods are proposed in GraphCut~\cite{grabcut}, Random Walk~\cite{randomWalkSegmentation}, Geodesic Matting~\cite{geodesicMattingSegmentation}, GSC~\cite{geodesicStartSegmentation} and ESC~\cite{geodesicStartSegmentation}.
On the other hand, deep-learning-based approaches using self-supervised learning techniques typically achieve significantly better results. 
Such methods rely on pre-trained models like DenseCL~\cite{DenseCL} or DINO~\cite{DINO} by utilizing network feature maps. 
Those methods include \cite{Simoni2021LocalizingOW}, \cite{Wang2022SelfSupervisedTF}, \cite{Shi00normalizedCuts}, \cite{hamilton2022unsupervised}, \cite{DSM}, \cite{FreeSOLO}, ~\cite{MIS} and~\cite{wang2024segment}.
Other methods are based on Stable Diffusion~\cite{StableDiffusion} (SD), where the attention maps are used to extract 
segments, including~\cite{DiffuseAttendSegment}, \cite{DiffusionDet} and \cite{paintSeg}.

Recently, M2N2~\cite{M2N2} was proposed, which significantly improved \sota NoC in the majority of considered datasets. However, M2N2 is sensitive to the attention map resolution, which poses a challenge to obtain low NoC's at 
high mIoU (e.g., at NoC95).
Inspired by M2N2~\cite{M2N2} and OIS~\cite{OISegmentation}, we integrate the depth map obtained from Depth Anything V2~\cite{DepthAnythingV2} into the Markov Map generation step and additionally introduce a segment size adaptation method. As a result, we obtain lower NoC values then in M2N2~\cite{M2N2} in all but the medical datasets, as well as lower NoC's then SAM~\cite{SAM} and SimpleClick~\cite{SimpleClick} in the DAVIS~\cite{DAVIS_DAtaset} and HQSeg44K~\cite{HQSAM} datasets.

%% file: sec/3_method.tex
\section{Method}\label{sec:method}
In this section we introduce M2N2V2 in full detail.
The whole architecture is illustrated in \cref{fig:framework_overview} with the novel contributions highlighted in blue.
As input, we are given an image $I \in \mathbb{R}^{H \times W \times 3}$ and a set of $N$ prompt points $X=\{ (x_1, y_1), (x_2, y_2), ..., (x_N, y_N) \}$ where each 2D prompt point $x_i \in \mathbb{R}^2$ has a class label $y_i \in \{0, 1\}$ with $0$ meaning that the prompt point belongs to the background and $1$ meaning that it belongs to the foreground.
Our framework can be separated into three stages.
First, we use the attention backbone to obtain a 4D attention tensor $\boldsymbol{A} \in \mathbb{R}^{h \times w \times h \times w}$ and use an additional depth backbone to generate a 2D full resolution depth map $D \in \mathbb{R}^{H \times W}$.
In the second stage, we fuse the low-resolution semantic information of $\boldsymbol{A}$ with the high resolution depth map to obtain depth-guided Markov-maps $M_{i} \in \mathbb{R}^{H \times W}$ for each prompt point $x_i$.
In the final stage, we scale each Markov-map $M_i$ to a scaled Markov-map $S_i \in \mathbb{R}^{H \times W}$ and combine all these maps in the truncated nearest neighbor to obtain the output segmentation $\hat{Y}_N \in \mathbb{Z}^{H \times W}$.
We improve the estimation of the optimal scale $\lambda_i^*$ of each scaled Markov-map $S_i$ by introducing a novel adaptive score function that takes into account the previous segmentation and the new prompt point.
The following subsections explain our approach in more detail.

\subsection{Attention Aggregation}
Similar to previous works \cite{DiffuseAttendSegment, M2N2}, we aggregate the multi-head self-attention tensors $\boldsymbol{F}_b \in \mathbb{R}^{N_{\mathrm{heads}} \times h \times w \times h \times w}$ of each attention block $b$ to create a single 4 dimensional tensor $\boldsymbol{A} \in \mathbb{R}^{h \times w \times h \times w}$.
This is done by computing the mean across all attention heads $N_\mathrm{heads}$ and calculating a weighted average over each attention block $b$:
\begin{align}
    \boldsymbol{A}[:, :, :, :] &= \sum_{b \in \mathrm{AttnBlocks}} w_{b} \sum_{n}^{N_{heads}} \boldsymbol{F}_b[n, :, :, :, :]
\end{align}
Here the sum of all block weights $w_b \ge 0$ is equal to $1$ to ensure that each attention map $A[i, j, :, :]$ represents a valid probability distribution.
We then adjust the temperature of the attention tensor and use iterative proportional fitting (IPF) to obtain a doubly stochastic tensor in which $\sum_{i,j} A[i, j, :, :] = \textbf{1}$, as well as $\sum_{i,j} A[:, :, i, j] = \textbf{1}$.

\subsection{Depth Map Generation}
We utilize the depth backbone to obtain a full resolution depth map $D \in \mathbb{R}^{H \times W}$ for a given input image $I$.
In our framework, we collect the inverse depth $\frac{1}{Z}$ to remove the bias of distant objects having significantly larger depth values than close ones.
We additionally normalize the depth map to a range of $[0, 1]$ to obtain the same range of values as the RGB channels used in the JBU step.

\subsection{Depth-guided Markov-maps}
M2N2 introduced the concept of Markov-iteration-maps, or short Markov-maps, to enhance semantic information present in the attention tensors.
The first step is to interpret the attention tensor $\boldsymbol{A}$ as a Markov transition operator.
This is done by reshaping the tensor $\boldsymbol{A}$ from $h \times w \times h \times w$ into a matrix $(h \cdot w) \times (h \cdot w)$.
Similar to a discrete-time Markov chain, given a start state $p_0 \in \mathbb{R}^{1 \times (h \cdot w)}$, the evolved state $p_t$ is calculated by repeated matrix multiplication:
\begin{align}
    p_t &= p_0 \cdot A^t
\end{align}
Here $p_0$ corresponds to the one-hot encoded position of the desired prompt point.
The key idea of Markov-maps is to measure for each pixel $k$ the number of iterations $t$ required to reach a relative probability threshold greater than $\tau \in (0, 1]$.
Using this process results in a vector $m \in \mathbb{Z}_{\ge 0}^{1 \times (h \cdot w)}$.
\begin{align}
    m[k] &= \min \left\{t\in \mathbb{Z}_{\ge 0}\ | \ \frac{p_t[k]}{\max p_t} > \tau \right\} \label{eqn:markov_map}
\end{align}
Reshaping the resulting vector $m$ into a matrix, a low resolution semantic Markov-map $M_\mathrm{sem} \in \mathbb{R}^{h \times w}$ is obtained.
The following steps involve using JBU for upscaling $M_\mathrm{sem}$ to image resolution and flood fill to enable instance segmentation by suppressing local minima of semantically similar regions.
We introduce depth as an additional guide in M2N2V2 with the following two approaches.
\\
\\
\textbf{Depth-guided JBU}. Our integration of depth information into the JBU processing step is simple and straightforward.
Since JBU already supports multi-channel inputs in the form of RGB images, we extend the 3 color channels with an additional depth channel.
We ensure that all four channels of the RGBD are in the same value range $[0, 1]$ to prevent any bias between the depth and color domain.
\\
\\
\textbf{Depth-guided Flood Fill}. Similar to the depth-guided JBU, we extend the flood fill approach to process two channels.
Instead of using the absolute distance between Markov-map values, we calculate the euclidean distance of a Markov-map and depth value pair.
After this step, we obtain the full resolution depth-guided Markov-maps $M \in \mathbb{R}^{H \times W}$.
We estimate a Markov-map $M_i$ for each individual prompt point $x_i$.

\begin{figure}[!ht]
  \centering
  \includegraphics[width=0.75\columnwidth]{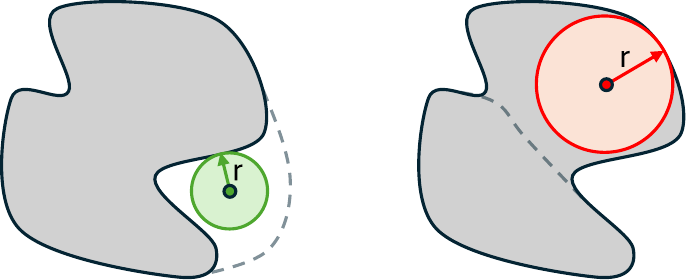}
  \caption{\textbf{Boundary distance estimation}. Two examples with the gray area representing the previous segmentation and the dashed line depicting the area the user wants to include (left) or exclude (right). On the left side, the user places a new foreground prompt point and on the right a background prompt point. The circle around each prompt point is drawn with a radius $r$ of the shortest distance to the previous segmentation boundary (shown as an arrow).}
  \label{fig:illustration_of_boundary_distance_circle}
\end{figure}

\definecolor{main_comparison_table_color}{rgb}{0.9, 0.9, 0.9}
\begin{table*}[h!]
  \centering
  \begin{tabular}{>{\kern-\tabcolsep}ll@{\hskip 5pt}c@{\hskip 3pt}c@{\hskip 12pt}c@{\hskip 3pt}c@{\hskip 12pt}c@{\hskip 3pt}c@{\hskip 12pt}c@{\hskip 3pt}c<{\kern-\tabcolsep}}
  
    \toprule
    \multirow{2}{*}{Method} & \multirow{2}{*}{Backbones} & \multicolumn{2}{@{}c@{\hskip 15pt}}{GrabCut~\cite{grabcut}} & \multicolumn{2}{@{}c@{\hskip 15pt}}{Berkeley~\cite{berkeleyDataset}} & \multicolumn{2}{@{}c@{\hskip 15pt}}{DAVIS~\cite{DAVIS_DAtaset}} & \multicolumn{2}{c}{HQSeg44K~\cite{HQSAM}} \\
    & & NoC90$\downarrow$ & NoC95$\downarrow$ & NoC90$\downarrow$ & NoC95$\downarrow$ & NoC90$\downarrow$ & NoC95$\downarrow$ & NoC90$\downarrow$ & NoC95$\downarrow$\\
    
    \midrule
    \rowcolor{main_comparison_table_color}
    \multicolumn{10}{l}{\textbf{Supervised}: Trained on Ground Truth Labels}\\
    SAM~\cite{SAM}                              & ViT-B & $1.84$ & $2.18$ & $2.06$ & $5.14$ & $5.14$ & $10.74$ & $\underline{7.46}$ & $12.42$ \\
    SimpleClick \cite{SimpleClick}              & ViT-B & $1.48$ & - & $1.97$ & - & $5.06$ & $10.37$ & $7.47$ & $\underline{12.39}$ \\
    SimpleClick                                 & ViT-H & $1.50$ & \bf{1.66} & $1.75$ & $4.34$ & $4.78$ & $8.88$ & - & - \\
    CPlot~\cite{clickPromptOptimalTransport}    & ViT-B & \bf{1.32} & - & $1.90$ & - & $4.65$ & - & - & - \\
    CFR-ICL~\cite{CFR-ICL}                      & ViT-H & $1.58$ & $\underline{1.76}$ & \bf{1.46} & \bf{2.90} & $\underline{4.24}$ & \bf{7.50}  & - & - \\
    FocalClick~\cite{FocalClick}                & SegFormerB3-S2 & $1.68$ & $1.92$ & $\underline{1.71}$ & $\underline{4.21}$ & $4.90$ & $10.40$ & - & - \\
    InterFormer-Tiny~\cite{interformer}         & ViT-L & $\underline{1.36}$ & - & $2.53$ & - & $5.21$ & - & - & - \\
    OIS~\cite{OISegmentation}                   & ViT-B & - & - & - & - & \bf{3.48} & $\underline{8.42}$ & \bf{3.47} & \bf{6.63}\\
    
    \midrule
    \midrule
    \rowcolor{main_comparison_table_color}
    \multicolumn{10}{l}{\textbf{Unsupervised}: Trained on Pseudo Labels}\\
    TokenCut$^\star$~\cite{TokenCutSO}          & ViT-B & $5.74$ & - & $9.97$ & - & $15.01$ & - & - & - \\
    FreeMask$^\star$~\cite{FreeMask}            & ViT-B & $6.10$ & - & $9.02$ & - & $13.05$ & - & - & - \\
    DSM$^\star$~\cite{DSM}                      & ViT-B & $4.64$ & - & $7.75$ & - & $10.11$ & - & - & - \\
    MIS~\cite{MIS}                              & ViT-B & $2.32$ & - & $4.58$ & - & $8.44$ & - & - & - \\

    \rowcolor{main_comparison_table_color}
    \multicolumn{10}{l}{\textbf{Unsupervised}: Training-Free}\\

    M2N2~\cite{M2N2}                                        & SD2 & $\underline{1.90}$ & $\underline{5.02}$ & $\underline{3.88}$ & $\underline{9.01}$ & $\underline{6.72}$ & $\underline{14.36}$ & $\underline{8.28}$ & $\underline{12.50}$\\
    \rowcolor{highlight} M2N2V2 (Ours)          & SD2, DA2 & \bf{1.84} & \bf{3.58} & \bf{2.25} & \bf{5.74} & \bf{4.77} & \bf{9.66} & \bf{6.29} & \bf{9.86}\\
    
    \bottomrule
  \end{tabular}
  \caption{\textbf{Comparison with previous work}. Methods indicated with $\star$ use the mentioned method to generate pseudo labels and then train SimpeClick~\cite{MIS} on these pseudo labels. All supervised methods are trained on the ground truth segmentation labels of SBD, except the InterFormer and SimpleClick (ViT-H), which are trained on COCO~\cite{COCO}+LVIS~\cite{LVIS} and OIS, which is trained on the HQSeg44K~\cite{HQSAM} training set.}
  \label{tab:main_comparison}
\end{table*}

\subsection{Adaptive Segmentation Size Score Function}
As in M2N2, we scale each depth-guided Markov-map $M_i$ by its individual divisor $\lambda_i^* \in \mathbb{R}_{>0}$ before applying the truncated nearest neighbor.
\begin{align}
    S_i = \frac{M_i}{\lambda_i^*}
\end{align}
Finding a good $\lambda_i^*$ is crucial, as all output pixels $x_q$ satisfying $S_i[x_q] > 1$ will be assigned the background class at the truncated nearest neighbor step.
To determine a good $\lambda_i^*$, M2N2 introduced a score function $s(\lambda, y_i, M_i, X)$ to evaluate each potential scale $\lambda$, where $\lambda_i^*$ maximizes the score.
\begin{align}
    \lambda_i^* = \;&\underset{\lambda}{\arg \max}~ s(\lambda, y_i, M_i, X)\\
    s(\lambda, y_i, M_i, X) = \;& 
    s_{\mathrm{prior}}(\lambda, M_i) \cdot 
    s_{\mathrm{edge}}(\lambda, M_i) \; \cdot \nonumber \\
    & s_{\mathrm{pos}}(\lambda, y_i, M_i, X) \cdot 
    s_{\mathrm{neg}}(\lambda, y_i, M_i, X) \label{eqn:total_score_function_definition}
\end{align}
As shown in \cref{eqn:total_score_function_definition}, the total score $s(\cdot)$ is composed of four individual score functions.
In order to prevent unreasonable segment size changes, we propose to replace $s_{\mathrm{prior}}(\lambda, M_i)$ (\cref{eqn:s-prior}) with $s_{\mathrm{adaptive}}(\lambda, M_i, x_i, y_i, \hat{Y}_{i-1})$ (\cref{eqn:s-adaptive}), where $\sigma_\mathrm{prior}$ is a constant threshold, $L_i :=L(x_i, y_i, \hat{Y}_{\mathrm{i-1}})$ is a function which estimates the maximum segment size change and $\Delta_i = |\mathcal{A}(\hat{Y}_i(\lambda)) - \mathcal{A}(\hat{Y}_{i-1})|$.
\begin{equation}\label{eqn:s-prior}
    s_\mathrm{prior}(\lambda, M_i) = \begin{cases}
        1 & \text{if } \mathcal{A}(\frac{M_i}{\lambda}) < \sigma_\mathrm{prior}  \cdot H\cdot W\\
        0 & \text{otherwise}
    \end{cases}
\end{equation}
\begin{align}
    s_\mathrm{adaptive}(\lambda, M_i, x_i, y_i, \hat{Y}_{i-1}) 
    = \begin{cases}
       1 & \text{if } \Delta_i < L_i\\ 
        0 & \text{otherwise}.
    \end{cases} \label{eqn:s-adaptive}
\end{align}
The function $\mathcal{A}(\cdot)$ returns the number of foreground pixels of a given segmentation $\hat{Y}$ or the number of pixels $\le 1$ for a scaled Markov-map $\frac{M_i}{\lambda}$.
Note that the segmentation $\hat{Y}_0$ is initialized as an empty segmentation with all pixels being background.
We hypothesize that prompt points close to the segment boundary should generally result in smaller segment size changes and we design the function $L$ accordingly.
\\
\\
\textbf{Controlling the segment size change}.
First, we define the function $r(x_i, y_i, \hat{Y}_{i-1})$ as the distance from $x_i$ to the previous segmentation $\hat{Y}_{i-1}$ boundary by calculating the Euclidean distance from $x_i$ to the closest same class pixel $x_q$ in $\hat{Y}_{i-1}$.
\begin{align}
    r(x_i, y_i, \hat{Y}_{i-1}) &= \min_{x_q} \begin{cases}
        ||x_q - x_i|| & \text{if } \hat{Y}_{i-1}[x_q] = y_i\\
        \infty & \text{otherwise}
    \end{cases} \label{eqn:boundary_distance_estimation}
\end{align}
\cref{fig:illustration_of_boundary_distance_circle} illustrates two examples of determining this distance.
Given $r$, we define $L$ as the area of a circle with the radius being the distance to the boundary $r$.
\begin{align}
    L(x_i, y_i, \hat{Y}_{\mathrm{i-1}}) &= \pi \cdot (\sigma_\mathrm{adaptive} \cdot r(x_i, y_i, \hat{Y}_{i-1}))^2 \label{eqn:heuristic}
\end{align}
The estimation of the maximal segment size change $L$ depends on the hyperparameter $\sigma_\mathrm{adaptive} \in \mathbb{R}_{>0}$.
While the contribution of the newest prompt point $x_N$ may not lead to excessive segment size change, previous prompts contributions still can do.
This is due to the fact that the scoring functions $s_\mathrm{pos}$ and $s_\mathrm{neg}$ both depend on the whole set of prompt points $X$.
Hence, we test if the new segment size change of $\hat{Y}_N$ exceeds the newest points limit $L(x_N, y_N, \hat{Y}_{\mathrm{N-1}})$.
If so, we discard the new prompt $x_N$ in all score function evaluations $s_{i=1,...,N-1}$.
This way, we ensure that the segment size change is definitely limited.

\subsection{Truncated Nearest Neighbor}
Following the design of M2N2, we fuse the scaled depth-guided Markov-maps $S_i$ of each prompt point $x_i$ and their corresponding class $y_i$ using a truncated nearest neighbor.
\begin{align}
    i^* &= \underset{i}{\arg \min}~ S_i[x_q]\label{eqn:nearest_neighbor}\\
    \hat{y}_q &= \begin{cases}
        y_{i^*} & \text{if } S_{i^*}[x_q] \le 1\\
        0 & \text{otherwise}
    \end{cases} \label{eqn:nearest_neighbor_turncation}
\end{align}
\cref{eqn:nearest_neighbor} determines the nearest neighbor prompt id $i^*$ and \cref{eqn:nearest_neighbor_turncation} estimates the output pixel class $\hat{y}_q \in \{0, 1\}$ of the output segmentation.
For a set of $N$ prompt points, we reshape and store the final segmentation in $\hat{Y}_N\in \mathbb{Z}^{H \times W}$.

\begin{figure*}[t!]
    \centering
    \begin{subfigure}[t]{0.46\textwidth}
        \centering
        \includegraphics[width=\columnwidth]{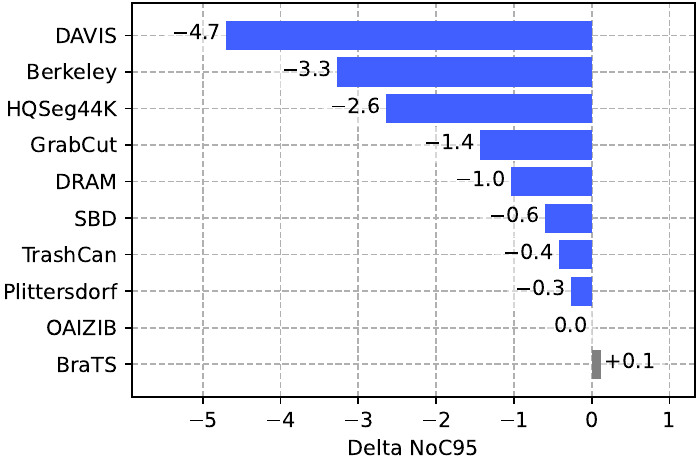}
    \end{subfigure}
    \hspace{0.079\textwidth}
    \begin{subfigure}[t]{0.46\textwidth}
        \centering
        \includegraphics[width=\columnwidth]{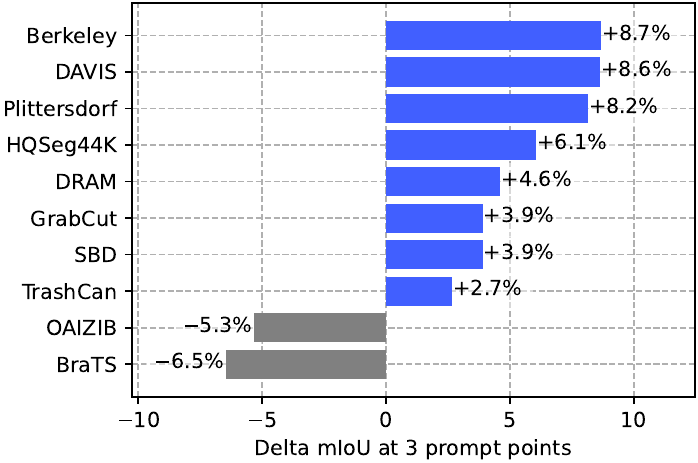}
    \end{subfigure}
    \caption{\textbf{M2N2V2 vs. M2N2}. Comparison of M2N2V2 with M2N2 on all 10 datasets, showing difference in NoC95 as well as mIoU@3.}
    \label{fig:delta_noc_and_iou}
\end{figure*}

\begin{figure*}
  \centering{\includegraphics[width=\textwidth]{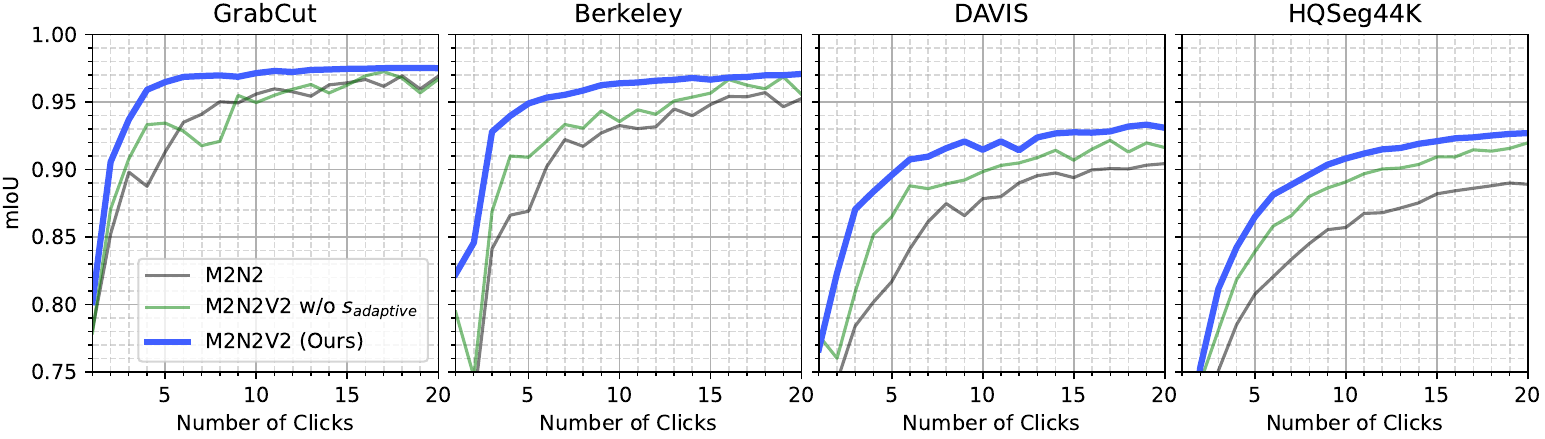}}
  \caption{\textbf{mIoU per NoC}. Using the adaptive segment size score function (blue curve) we achieve a higher and smoother mIoU curve.}
  \label{fig:mIoU_per_NoC}
\end{figure*}

%% file: sec/4_experiments.tex
\section{Experiments}\label{sec:experiments}
\textbf{Datasets}. We perform experiments on a total of \textbf{10} publicly available datasets. 
Our main comparisons are on the high segmentation quality datasets DAVIS~\cite{DAVIS_DAtaset} and HQSeg44K~\cite{HQSAM}.
We additionally perform experiments on GrabCut~\cite{grabcut}, Berkeley~\cite{berkeleyDataset}, SBD~\cite{SBD-Dataset}, Plittersdorf~\cite{plittersdorf}, TrashCan~\cite{trashcan-dataset}, 2 datasets from the medical domain called OAI-ZIB~\cite{OAIZIB} and BRATS~\cite{BraTS}, and a dataset of paintings called DRAM~\cite{dram-dataset}.
Please see \cref{sec:datasets} for a description of each dataset.
\\
\\
\textbf{Evaluation metrics}.
Following previous work \cite{SimpleClick, MIS, M2N2}, we simulate the user interaction with our model by placing the next click in the center of the largest error region.
We provide two metrics for the average number of clicks required to reach an IoU of $90\%$ as NoC90 and an IoU of $95\%$ as NoC95, with a maximum number of clicks per instance of 20.
We also measure the mean IoU which is reached after N clicks as mIoU@N.
\\
\\
\textbf{Implementation details}.
We use the same set of hyperparameters on all datasets with Stable Diffusion 2 (SD2) as an attention backbone and Depth Anything V2 (DA2) for the depth estimation.
We configure all hyperparameters as in M2N2 except for $\sigma_\mathrm{adaptive}=6$.
To reduce the seconds per click (SPC), we cache the attention tensor, depth map, and previously calculated Markov-maps, respectively.
On the DAVIS dataset, with an image resolution of $854 \times 480$, we require $0.7$ seconds per attention tensor and $1.7$ seconds per depth map on a single GeForce RTX 4090.
Without including the pre-loading of attention and depth, we achieve an SPC of $0.48$.
We provide further implementation details in \cref{sec:futher_implementation_details} together with a list of all considered backbones in \cref{sec:backbones_in_detail}.

\begin{table*}[h!]
  \centering
  \begin{tabular}{>{\kern-\tabcolsep}l@{\hskip 8pt}l@{\hskip 15pt}c@{\hskip 4pt}c@{\hskip 4pt}c@{\hskip 4pt}c@{\hskip 15pt}c@{\hskip 4pt}c@{\hskip 4pt}c@{\hskip 4pt}c<{\kern-\tabcolsep}}
  
    \toprule
    \multicolumn{2}{@{}c@{\hskip 15pt}}{Backbones} & \multicolumn{4}{@{}c@{\hskip 15pt}}{DAVIS~\cite{DAVIS_DAtaset}} & \multicolumn{4}{c}{HQSeg44K~\cite{HQSAM}} \\
    Attention & Depth & NoC90$\downarrow$ & NoC95$\downarrow$ & mIoU@5$\uparrow$ & mIoU@10$\uparrow$ & NoC90$\downarrow$ & NoC95$\downarrow$ & mIoU@5$\uparrow$ & mIoU@10$\uparrow$\\
    
    \midrule

		SD2 & - & $6.43$ & $14.3$ & $0.868$ & $0.896$ & $8.24$ & $12.5$ & $0.823$ & $0.869$\\
		SD2 & Marigold & $5.58$ & $13.5$ & $0.877$ & $0.892$ & $7.75$ & $12.4$ & $0.833$ & $0.879$\\
		- & DA2 & $14.5$ & $17.8$ & $0.664$ & $0.768$ & $9.75$ & $12.8$ & $0.758$ & $0.827$\\
		ViT-B & DA2 & $5.91$ & $11.7$ & $0.882$ & $0.910$ & $6.92$ & $10.5$ & $0.856$ & $0.897$\\
		SD1 & DA2 & \underline{5.03} & \underline{10.4} & \underline{0.887} & \underline{0.912} & \textbf{6.24} & \textbf{9.84} & \textbf{0.867} & \textbf{0.909}\\
		SD2 & DA2 & \textbf{4.76} & \textbf{9.66} & \textbf{0.896} & \textbf{0.915} & \underline{6.29} & \underline{9.86} & \underline{0.865} & \underline{0.908}\\
    
    \bottomrule
  \end{tabular}
  \caption{\textbf{Backbone Comparison}. Evaluating the performance impact of choosing different backbones for M2N2V2. We show the NoC and IoU results of the DAVIS and HQSeg44K dataset.}
  \label{tab:backbones_comparison}
\end{table*}

\begin{table*}[h!]
  \centering
  \begin{tabular}{>{\kern-\tabcolsep}l@{\hskip 10pt}c@{\hskip 4pt}c@{\hskip 4pt}c@{\hskip 4pt}c@{\hskip 15pt}c@{\hskip 4pt}c@{\hskip 4pt}c@{\hskip 4pt}c<{\kern-\tabcolsep}}
  
    \toprule
    \multirow{2}{*}{Version} & \multicolumn{4}{@{}c@{\hskip 15pt}}{DAVIS~\cite{DAVIS_DAtaset}} & \multicolumn{4}{c}{HQSeg44K~\cite{HQSAM}} \\
     & NoC90$\downarrow$ & NoC95$\downarrow$ & mIoU@5$\uparrow$ & mIoU@10$\uparrow$ & NoC90$\downarrow$ & NoC95$\downarrow$ & mIoU@5$\uparrow$ & mIoU@10$\uparrow$\\
    
    \midrule

		w/o Novel Components (M2N2) & $6.72$ & $14.4$ & $0.817$ & $0.878$ & $8.28$ & $12.5$ & $0.808$ & $0.857$\\
		w/o Depth Guided Markov-maps & $6.43$ & $14.3$ & $0.868$ & $0.896$ & $8.24$ & $12.5$ & $0.823$ & $0.869$\\
		w/o Depth Guided JBU & $4.93$ & $10.7$ & \underline{0.892} & \underline{0.912} & $6.46$ & $10.1$ & \underline{0.861} & \underline{0.904}\\
		w/o Depth Guided Flood Fill & $5.47$ & $12.2$ & $0.887$ & $0.903$ & $7.59$ & $11.5$ & $0.834$ & $0.88$\\
		w/o Adaptive Score Function & \underline{4.81} & \underline{9.71} & $0.865$ & $0.898$ & \textbf{6.27} & \textbf{9.85} & $0.839$ & $0.891$\\
		Full M2N2V2 Framework & \textbf{4.76} & \textbf{9.66} & \textbf{0.896} & \textbf{0.915} & \underline{6.29} & \underline{9.86} & \textbf{0.865} & \textbf{0.908}\\
    
    \bottomrule
  \end{tabular}
  \caption{\textbf{Impact of M2N2V2's novel components}. Showing the impact each novel component in the M2N2V2 framework. We show results on both NoC and mIoU for the DAVIS and HQSeg44K dataset.}
  \label{tab:impact_of_novel_components}
\end{table*}

\subsection{Comparison with Previous Work}
\cref{tab:main_comparison} shows the comparison of our method with previous work.
We split the table into three categories of supervised, unsupervised with training on pseudo labels, and unsupervised without any training.
Our method does not require any ground truth labels nor additional training and therefore belongs to the third category.
We want to highlight that the comparison between categories is not fair, as the task of training-free and unsupervised methods is significantly more challenging due to missing ground truth labels.
Our model M2N2V2 significantly outperforms the previous state-of-the-art M2N2.
In the NoC95 category we observe a reduction of about $2.5$ clicks in the HQSeg44K dataset.
Interestingly, we achieve competitive results compared to supervised methods such as SAM or SimpleClick \textbf{without any training on segmentation labels}.
Furthermore, we notice that our method requires only about $1.3$ more clicks on the DAVIS dataset compared to the supervised state-of-the-art OIS model.
This difference increases for the HQSeg44k dataset, in which we require approximately $3$ additional clicks on average, but it is important to note that OIS is trained on the HQSeg44k training split and might have a domain advantage.
Comparing M2N2V2 and M2N2 in \cref{fig:delta_noc_and_iou} directly, we observe clear improvements in both NoC95 as well as the mIoU@3 on 8 of the 10 evaluation dataset.
This also includes paintings in the DRAM dataset, grayscale images of wildlife in the Plittersdorf dataset, or even underwater trash detection in the TrashCan dataset.
In the cases of the 2 medical datasets OAIZIB and BraTS, we notice that the NoC and mIoU are not improving, which is probably because the depth map does not provide any useful information in the medial domain.
We also show the mIoU per click on the GrabCut, Berkeley, DAVIS and HQSeg44K datasets in \cref{fig:mIoU_per_NoC}, in which M2N2V2 achieves consistently higher mIoU than M2N2.

\begin{figure*}
  \centering{\includegraphics[width=\textwidth]{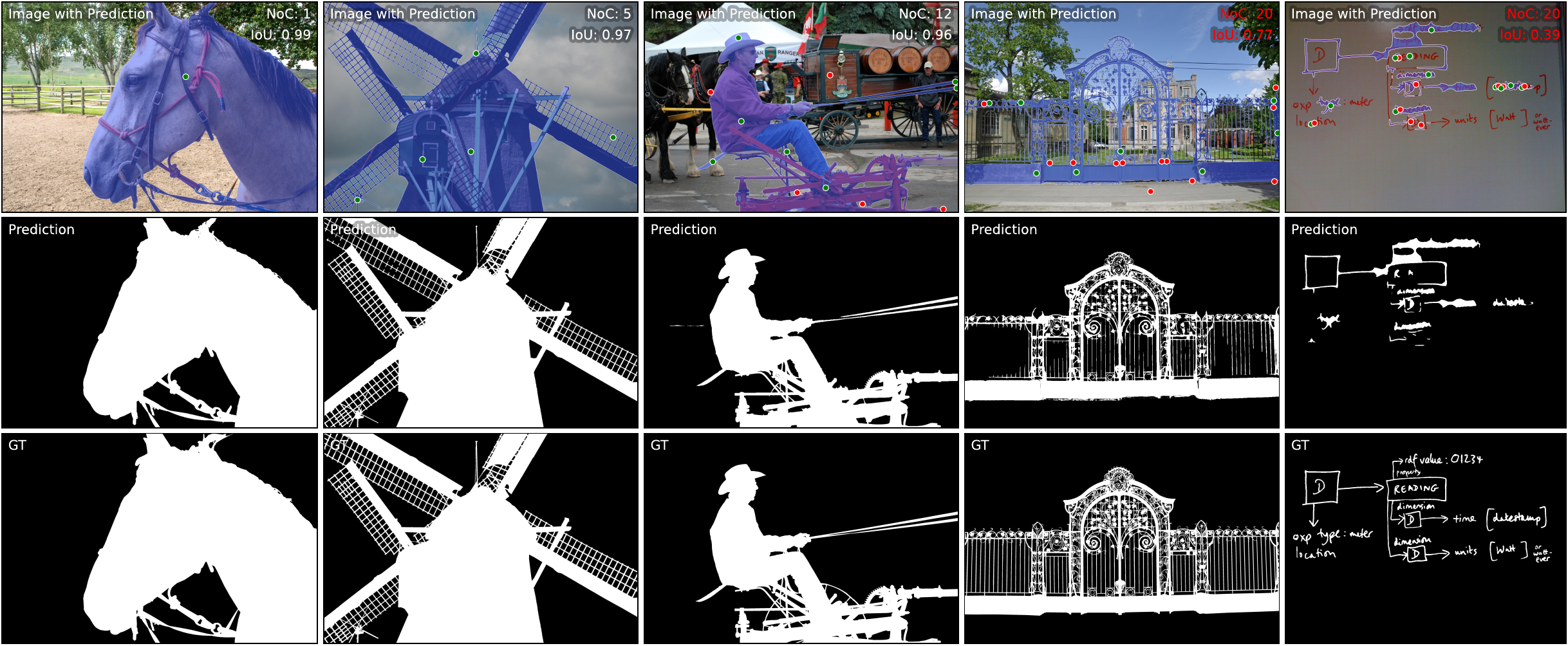}}
  \caption{\textbf{Example predictions on the HQSeg44K Dataset}. Easy to more difficult images are sorted from left to right. Foreground prompt points are shown in green and background prompt points in red. M2N2V2 is able to segment the majority of fine structures successfully.}
  \label{fig:qualitative_examples}
\end{figure*}

\subsection{Ablation Study}
\textbf{Comparison of backbones}.
M2N2V2's design allows easy drop-in of various backbones.
We therefore perform various experiments on multiple backbones and present the results in \cref{tab:backbones_comparison}.
We note that SD1 and SD2 with DA2 show the best results, SD1 having slightly better NoC on the HQSeg44K, while SD2 has better NoC on DAVIS by a larger margin.
We observe that excluding attention maps and only using depth (DA2) yields worst results. This indicates that the attention contributes the most important semantic information.
Furthermore, using depth is consistently beneficial as it improves the results in all considered metrics. We provide a full description of all considered backbones in \cref{sec:backbones_in_detail}.
\\
\\
\textbf{Depth Map Guidance}.
In \cref{tab:impact_of_novel_components} we can observe the performance impact of the proposed depth-guided JBU and depth-guided flood fill within the Markov-map generation process. We observe that depth-guided flood fill contributes the most to the performance.  
Discarding depth entirely (w/o Depth Guided Markov-maps) brings the NoC almost to the same performance as M2N2 (first row).
We also investigated integrating the depth map into the Markov-process without any significant improvements (see \cref{sec:depth_aware_attention}).
Looking at the qualitative examples provided in \cref{fig:qualitative_examples}, we observe that M2N2V2 is able to segment fine details with high quality.
This especially applies to segments having a significant contrast in their foreground/background depth values.
M2N2 struggles in such cases, as the attention maps only provide semantic information and the flood fill approach only suppresses non-overlapping instances.
If the overlapping instances are of different distance to the camera, M2N2V2 is able to separate them due to their different depth values.
We provide more qualitative examples in \cref{sec:depth_guidance_in_detail}.
\\
\\
\textbf{Adaptive Score Function}.
\cref{tab:impact_of_novel_components} shows the impact of the adaptive score function.
We can see that the impact on the NoC is not as significant as it is on the mIoU.
This behavior is expected, as the main goal is to prevent large sudden segment size changes which often occur after the desired IoU has already been reached.
\cref{fig:mIoU_per_NoC} shows the improvement on mIoU more clearly.
M2N2V2 with the $s_\mathrm{adaptive}$ component achieves higher mIoU values with less clicks on all four datasets.
Furthermore, the graph of M2N2V2 with $s_\mathrm{adaptive}$ is smoother on the GrabCut and Berkeley datasets, as preventing unreasonable changes in segment size prevents sudden collapses of the IoU.
We provide two interaction examples in \cref{sec:interactive_examples}, showing how applying $s_\mathrm{adaptive}$ prevents such cases.
Additionally, in \cref{sec:random_vs_center_point_sampling}, we show that M2N2V2 also benefits from the proposed adaptive score function in different point sampling strategies.

\subsection{Limitations}
M2N2V2 performs worse in cases where the depth map does not provide useful information.
For example, we can observe this in the two medical datasets, as shown in \cref{fig:delta_noc_and_iou}.
Similarly, in the rightmost column in \cref{fig:qualitative_examples}, the target segmentation is on a flat surface. As a result, M2N2V2 has difficulties segmenting the red text and only achieves an IoU of $39\%$ after 20 points.
Another reason for the difficulty in this case is that the desired segmentation consists of multiple disconnected regions, as flood fill suppresses local minima in the Markov-maps and therefore allows only a connected segment region per prompt point. Note that it is still possible to segment disconnected regions by providing multiple prompt points.
In the majority of cases this is not an issue, as single instances are usually a connected region in the image.
The same challenge occurs in obstructing views, as they may lead to disconnected segment regions as well.
Similar to M2N2, fine structures can still pose a challenge if the depth guidance in M2N2V2 does not provide separation support (e.g., due to insufficient depth contrast), as can be seen in \cref{fig:failure_cases_qualitative_examples}.

%% file: sec/5_conclusion.tex
\section{Conclusion}\label{sec:conclusion}
We introduced M2N2V2, a novel method to utilize both attention and depth maps without any additional training to enable unsupervised point prompt-based interactive segmentation.
In extensive experiments we showed that the addition of high resolution depth maps as another modality combined with segment size adaptation significantly improves the NoC and the mIoU values, underlying the benefits of multi-modal approaches especially on unsupervised settings.
Future work may involve introducing instance information via cross attention or supporting different type of prompts like text, box or scribble.

%% file: sec/6_supplementary.tex
\clearpage
\setcounter{page}{1}
\maketitlesupplementary

\begin{table*}[t!]
    \setlength\extrarowheight{2pt} 
    \centering
    \begin{tabularx}{\textwidth}{|c|X|c|c|c|c}
        \hline
        {\bf Dataset} & {\bf Description} & {\bf Image Type} & {\bf Images} & {\bf Masks} \\
        \hline
        GrabCut~\cite{grabcut} & Different kinds of foreground objects, e.g.,  people, vehicles, animals, etc. Initially created to test the GrabCut~\cite{grabcut} method & Photos & $50$ & $50$\\
        \hline
        Berkeley~\cite{berkeleyDataset} & Primary Purpose of this dataset is to provide standardized data for training and evaluating segmentation and edge-detection algorithms. A subset of the images is also found in the GrabCut dataset. & Photos & $96$ & $100$\\
        \hline
        DAVIS~\cite{DAVIS_DAtaset} & A Dataset for video object segmentation, where objects might be occluded and comprise motion blur. Images are taken from video frames and we use the same 345 frames as used in \cite{SimpleClick, MIS}. & Videos & $345$ & $345$\\
        \hline
        HQSeg44K~\cite{HQSAM} & A collection of multiple high quality datasets containing high resolution images. We use the same validation subset as \mbox{SegNext}~\cite{SegNext-InteractiveSegmentation}. & Photos & $1537$ & $1537$\\
        \hline
        SBD~\cite{SBD-Dataset} & The Semantic Boundaries Dataset (SBD) is a dataset for predicting pixels on the boundary of the object. 
        & Photos & $2857$ & $4119$\\
        \hline
        Plittersdorf~\cite{plittersdorf} & This dataset was captured using the SOCRATES stereo camera trap in early 2022 in the wildlife park Plittersdorf, Bonn Germany. It includes instance segmentation of wildlife on gray-scale images. & Wildlife & $50$ & $151$\\
        \hline
        TrashCan~\cite{trashcan-dataset} & The TrashCan dataset is an instance-segmentation dataset of underwater trash. & Underwater & $1147$ & $2588$\\
        \hline
        DRAM~\cite{dram-dataset} & Semantic segmentation masks of art. We predict every mask except the background. & Art & $718$ & $1179$\\
        \hline
        OAIZIB~\cite{OAIZIB} & Dataset for segmentation of knee bone and cartilage from MRI scans. & Medical & $150$ & $150$\\
        \hline
        BraTS~\cite{BraTS} & Dataset for brain tumor segmentation based on MRI Scans.  & Medical & $369$ & $369$\\
        \hline
    \end{tabularx}
    \caption{\textbf{List of all Datasets}.}
    \label{tab:datasets}
\end{table*}

\begin{table*}[t!]
    \setlength\extrarowheight{2pt} 
    \centering
    \begin{tabularx}{\textwidth}{|l|X|c|l|}
        \hline
        {\bf Backbone} & {\bf Full Name and Description} & {\bf Backbone Type} & {\bf Hugging Face Model Id}\\
        \hline
        ViT-B~\cite{vit} & Self supervised pre-trained \textbf{Vision Transformer}. We use the DinoV2 pre-trained weights and collect only the attention of the last attention layer. & Attention & {\tiny facebook/dinov2-base}\\
        \hline 
        SD1~\cite{StableDiffusion} & \textbf{Stable Diffusion 1.5}. Text to image diffusion model. The architecture is a UNet with transformer blocks. We aggregate the same 2 self-attention layers with equal weight as M2N2~\cite{M2N2} & Attention & {\tiny CompVis/stable-diffusion-v1-1}\\
        \hline
        SD2~\cite{StableDiffusion} & \textbf{Stable Diffusion 2.0}. A new version of the SD1 model with the same architecture but training on a higher image resolution.  We aggregate the same 2 self-attention layers with equal weight as M2N2~\cite{M2N2} & Attention & {\tiny stabilityai/stable-diffusion-2}\\
        \hline
        DA2~\cite{DepthAnythingV2} & \textbf{Depth Anything V2}. We use the largest model available during drafting of this paper, called "Large". As suggested in the paper, we upscale the image first to obtain a more detailed result. We choose a size of 2024 pixels for the shortest side of the input image and then resize the output back to the image resolution. & Depth & {\tiny depth-anything/Depth-Anything-V2-Large-hf}\\
        \hline
        Marigold~\cite{Repurposing_SD_for_MonoDepth} & \textbf{Marigold}. As advised by the authors, we run the model with 50 inference steps and an ensemble size of 10. We then convert the output to the inverse depth by $\frac{1}{Z+0.25}$. & Depth & {\tiny prs-eth/marigold-depth-lcm-v1-0}\\
        \hline
    \end{tabularx}
    \caption{\textbf{List of all Backbones}.}
    \label{tab:backbones}
\end{table*}

\begin{table}[t!]
    \centering
    \begin{tabular}{@{}ccccc@{}}
        \toprule
        $\sigma_{\mathrm{adaptive}}$ & NoC90$\downarrow$ & NoC95$\downarrow$ & mIoU@5$\uparrow$ & mIoU@10$\uparrow$ \\
        \midrule
        $2$ & $5.65$ & $11.8$ & $0.894$ & $0.909$\\
		$3$ & $4.91$ & $10.6$ & \underline{0.896} & $0.913$\\
		$4$ & \textbf{4.74} & $10.3$ & \textbf{0.902} & \textbf{0.917}\\
		$5$ & \underline{4.76} & $9.90$ & \textbf{0.902} & \textbf{0.917}\\
		$6$ & \underline{4.76} & \textbf{9.66} & \underline{0.896} & \underline{0.915}\\
		$7$ & $4.77$ & \underline{9.77} & $0.894$ & $0.916$\\
        \bottomrule
    \end{tabular}
    \caption{\textbf{Comparison of $\sigma_{\mathrm{adaptive}}$ on DAVIS}. Using $s_\mathrm{adaptive}$ results in an improvement in mIoU independent of the prompt point sampling method.}
    \label{tab:adaptive_heuristic_noc_miou_tradeoff}
\end{table}

\begin{figure}[!ht]
  \centering
  \includegraphics[width=\columnwidth]{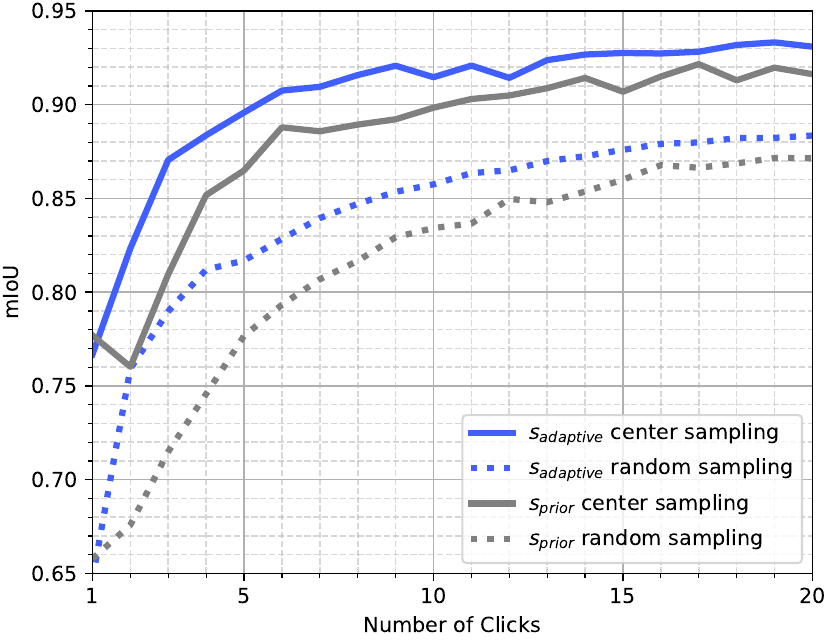}
  \caption{\textbf{Center Sampling vs. Random Sampling}. Comparing the effect of our proposed score function for segment size change control. We use two point sampling strategies for the simulated user interaction on the DAVIS dataset. Both graphs of random sampling are averaged values over 5 runs. We can see that this score function is effective on both sampling strategies.}
  \label{fig:random_sampling_vs_center_sampling}
\end{figure}

\begin{figure*}
  \centering{\includegraphics[width=\textwidth]{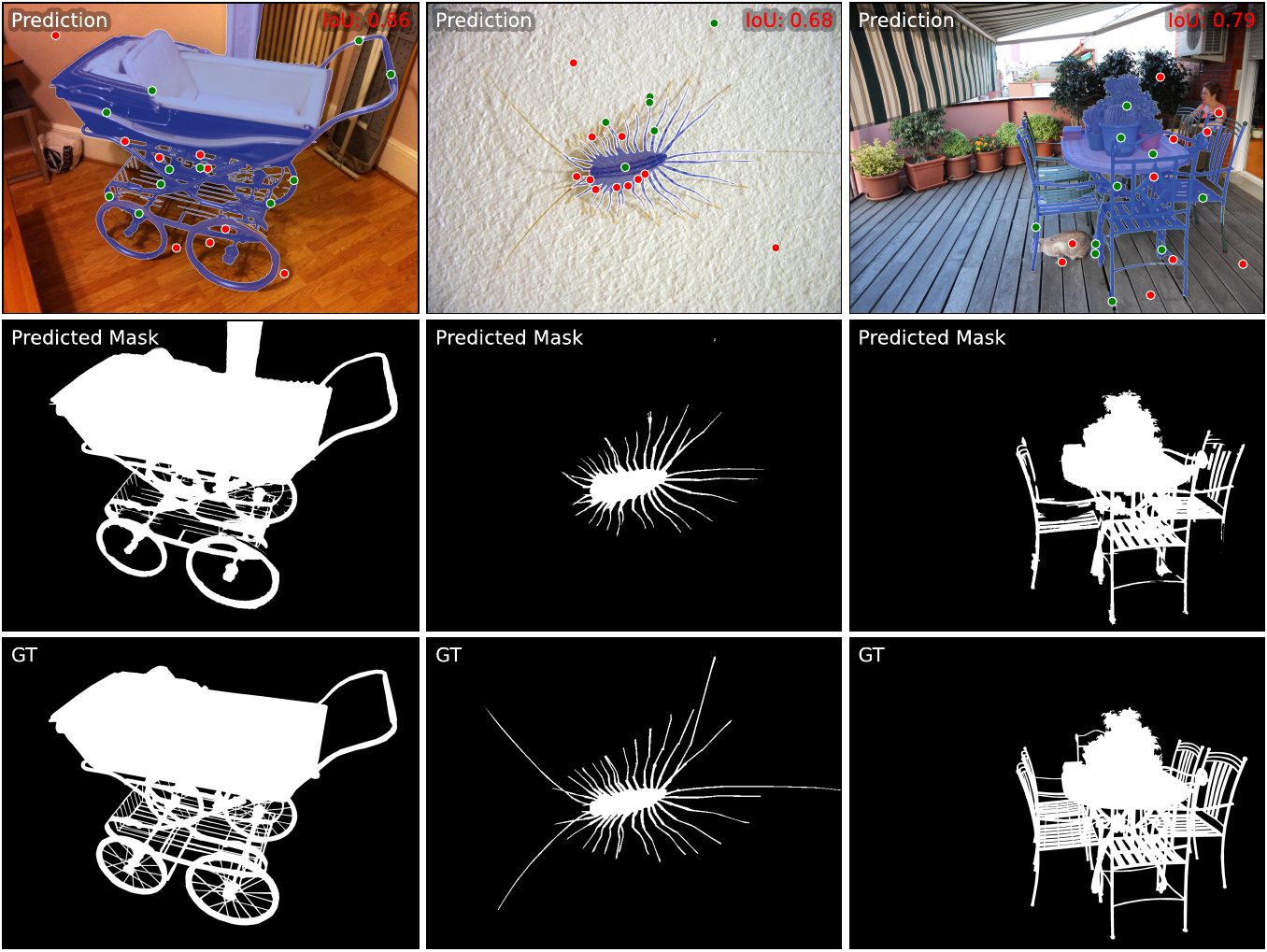}}
  \caption{\textbf{M2N2V2 more failure cases in HQSeg44K}. The examples show that while our method is able to segment finer structures, it is still not able to  }
  \label{fig:failure_cases_qualitative_examples}
\end{figure*}

\section{Datasets}\label{sec:datasets}
We provide a full list of all 10 datasets in \cref{tab:datasets}.
Except for DRAM~\cite{dram-dataset}, which only provides semantic labels, we use the instance labels for evaluation in all datasets. In case a dataset has no test set we use the validation set.

\section{Backbones} \label{sec:backbones_in_detail}
All backbones are downloaded and used with the Hugging Face diffusers package.
Please see \cref{tab:backbones} for a detailed list of all backbones.
We select the attention layer aggregation weights following \cite{M2N2}.

\section{Adaptive Score Hyperparameter}
We choose the same $\sigma_\mathrm{adaptive}=6$ for all evaluations.
A comparison of different values for $\sigma_\mathrm{adaptive}$ is shown in \cref{tab:adaptive_heuristic_noc_miou_tradeoff}.
Notably, reducing the $\sigma_\mathrm{adaptive}$ to values below $4$ has little impact on the mIoU but increases NoC significantly.
This is due to the fact that discarding larger changes in segment size can result in slower, but more controlled, segment growth.
We therefore choose $\sigma_\mathrm{adaptive}=6$ to get the benefit of a controlled segment size growth without significantly reducing the NoC. Note that M2N2V2 has the same amount of hyperparameters as M2N2.

\section{Further Implementation Details}\label{sec:futher_implementation_details}
The entire framework is implemented in python and PyTorch.
To limit memory usage, we keep the the shortest side of an input image to a maximum of $1024$ pixels for both M2N2V2 and M2N2.
We achieve this by resizing images which exceed this size while keeping their aspect ratio the same.
This only has an impact on the HQSeg44K dataset, as all other datasets do not have smaller image resolutions.
We use additional test time augmentation in the depth and attention tensor aggregation.
We do this by flipping the image horizontally and average pairs of attention tensors and depth maps.
Please note that the main target of our paper is to investigate the potential of unsupervised and training-free interactive segmentation and we therefore have not fully optimized the code.
The flood fill approach is still implemented on the CPU using just in time compilation and numpy arrays to speed up the computation.

\section{Depth Guided Attention} \label{sec:depth_aware_attention}
OIS~\cite{OISegmentation} introduces a relative depth map (order map) per prompt point to the attention, by adding it to the key query product, which is named as order-aware attention.
We also investigated adding the depth information as a post processing step, similar to Iterative Proportional Fitting (IPF) or adjusting the temperature.
In our experiments within the M2N2V2 framework, using order-aware attention as post processing step mainly increased the NoC and only improved performance in rare cases.
There are multiple possible reasons for this.
One reason might be artifacts that are created, when we downscale the depth map from image resolution $H \times W$ to attention map resolution $h \times w$.
Another reason might be that the depth-aware attention is only helpful within a Neural Network and its training process, as in OIS~\cite{OISegmentation}.
Finally, the depth information might suppress the semantic information present in the attention maps and therefore decrease the segmentation performance.
In the end, we decided not to pursue depth-aware attention, as we already introduce the depth information in depth guided JBU and depth aware flood fill which both proved to be significantly more effective in reducing the NoC.

\section{Dependence of Adaptive Score Function on Prompt Point Sampling}\label{sec:random_vs_center_point_sampling}
The adaptive score function $s_\mathrm{adaptive}$ depends on the distance $r$ to the segment boundary.
Since we use the center of the error region to place the next prompt point in our evaluation, there might be a slight advantage in knowing that the prompt point will only be close to the segment boundary if the error region is small.
We therefore perform an additional evaluation of M2N2V2 on the DAVIS dataset with a random point prompt sampling strategy.
In the random sampling strategy, each new prompt point is sampled uniformly over the error region.
This effectively means that points can be close to the boundary of the previous region, even if the error region is large.
We show the impact of choosing random sampling in \cref{fig:random_sampling_vs_center_sampling}.
We notice that the novel scoring function $s_\mathrm{adaptive}$ outperforms the previous scoring function $s_\mathrm{prior}$ independent of the choice of the prompt point sampling method.

\begin{figure*}
  \centering{\includegraphics[width=\textwidth]{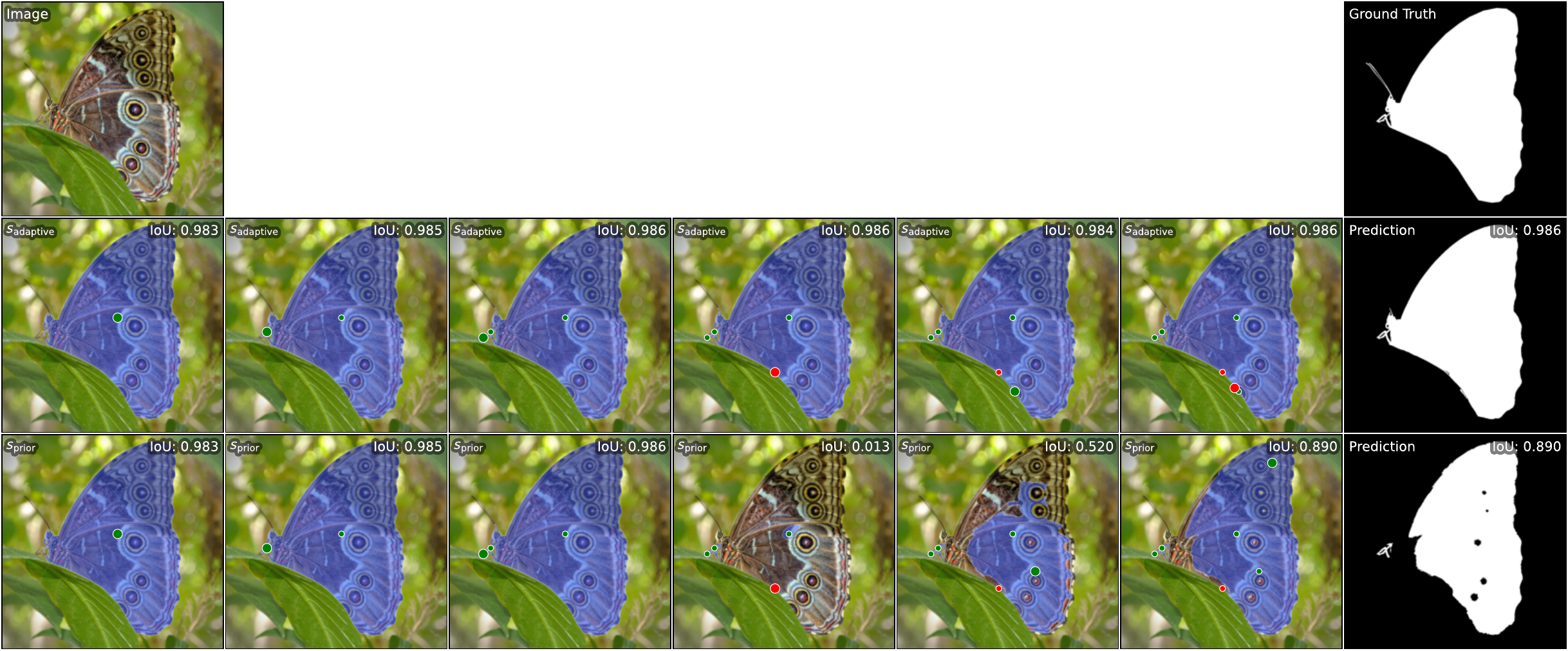}}
  \caption{\textbf{Prevention of Undesired Large Area Exclusion}. In this example, the third point (4th row, red background point) causes the segmentation to collapse in the last row with $s_\mathrm{prior}$. It takes additional prompt points to correct this collapse but after 2 more points (column 6) the segmentation has still not yet recovered from the red point. The novel score function $s_\mathrm{adaptive}$ prevents this from happening.}
  \label{fig:interactive_collapse_to_small}
\end{figure*}

\begin{figure*}
  \centering{\includegraphics[width=\textwidth]{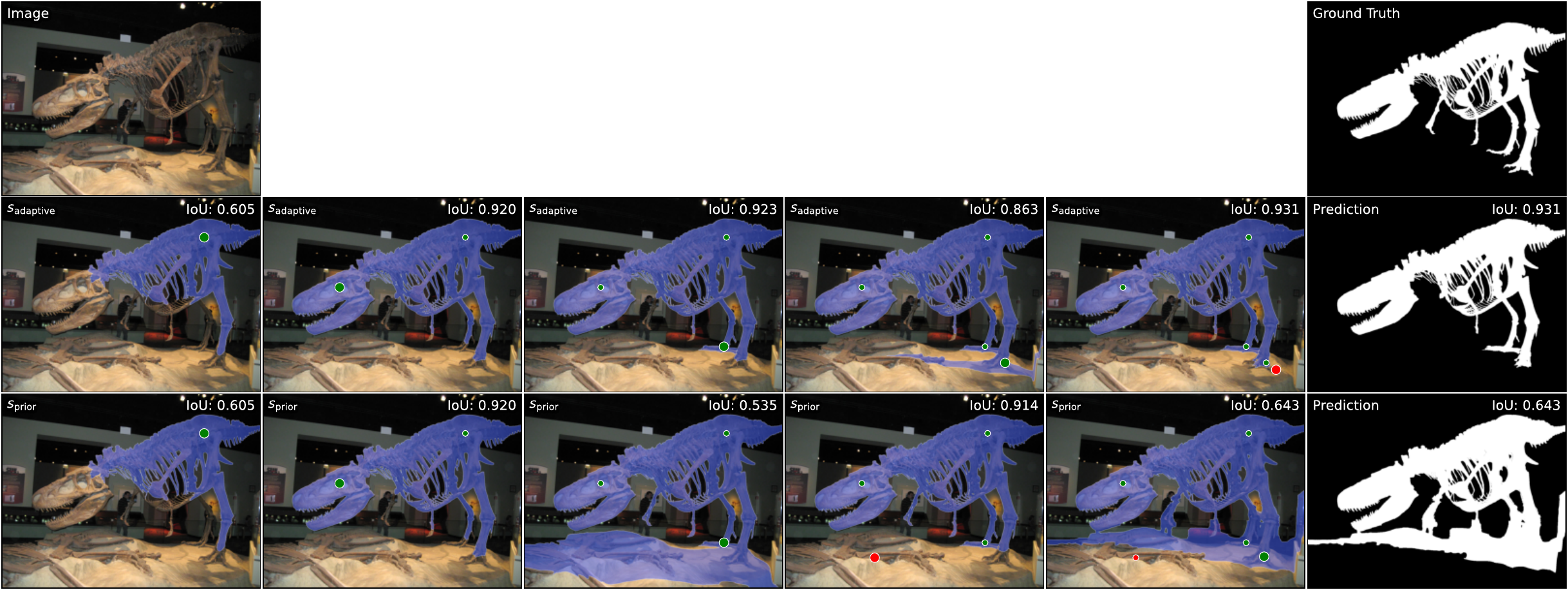}}
  \caption{\textbf{Prevention of Undesired Large Area Inclusion}. In this example the segmentation in last row using $s_\mathrm{prior}$ sometimes includes undesired large regions of the floor. This happens because a foreground point is placed on the boundary of the dinosaur and the floor. The $s_\mathrm{adaptive}$ reduces the affect of the these points almost entirely, but in column 4 we still observe a small undesired strip of the floor included in the segmentation.}
  \label{fig:interactive_expand_to_big}
\end{figure*}

\section{Interactive Examples of the Adaptive Score Function}\label{sec:interactive_examples}
We provide two examples showing the impact of the adaptive score function.
First, \cref{fig:interactive_collapse_to_small} shows the example of segmenting a butterfly.
Using M2N2's $s_\mathrm{prior}$, a background point which is set too close to the segment boundary causes the segmentation to collapse, down from a previously IoU of $99\%$ to $1\%$.
This is due to the prompt point being too close to the butterfly which leads to the exclusion of the entire butterfly wing.
This faulty segmentation requires multiple corrective points to include the wing back into the segment again.
In the provided example, the two points are not enough to recover the segmentation, resulting in an IoU of $89\%$ at 6 prompt points compared to $99\%$ at only 3 points.
M2N2V2's $s_\mathrm{adaptive}$ successfully prevents this collapse and keeps an IoU of $99\%$ at 6 prompt points.
The second example in \cref{fig:interactive_expand_to_big} shows a foreground point causing a too large segmentation including background regions.
This can be noticed at the $s_\mathrm{prior}$ based segmentation, where undesired background regions are falsely included in the segmentation at the 3rd and 5th points, respectively.
Similar to the previous example, $s_\mathrm{adaptive}$ also prevents this large change of the segment size, as the prompt point is very close to the boundary of the previous segmentation.
Nonetheless, we notice that the 4th point still results in a small region of the floor included even with $s_\mathrm{adaptive}$.
It is important to note that some users might want to segment the floor together with the dinosaur.
This is indeed still possible with M2N2V2, we only constrain the segment size if the point is placed very closely to the previous segmentation boundary.
For the interaction in \cref{fig:interactive_expand_to_big}, the points are all placed very close to the dinosaurs foot, we therefore can expect that the user only wants to segment these and not the entire floor.

\begin{figure*}
  \centering{\includegraphics[width=\textwidth]{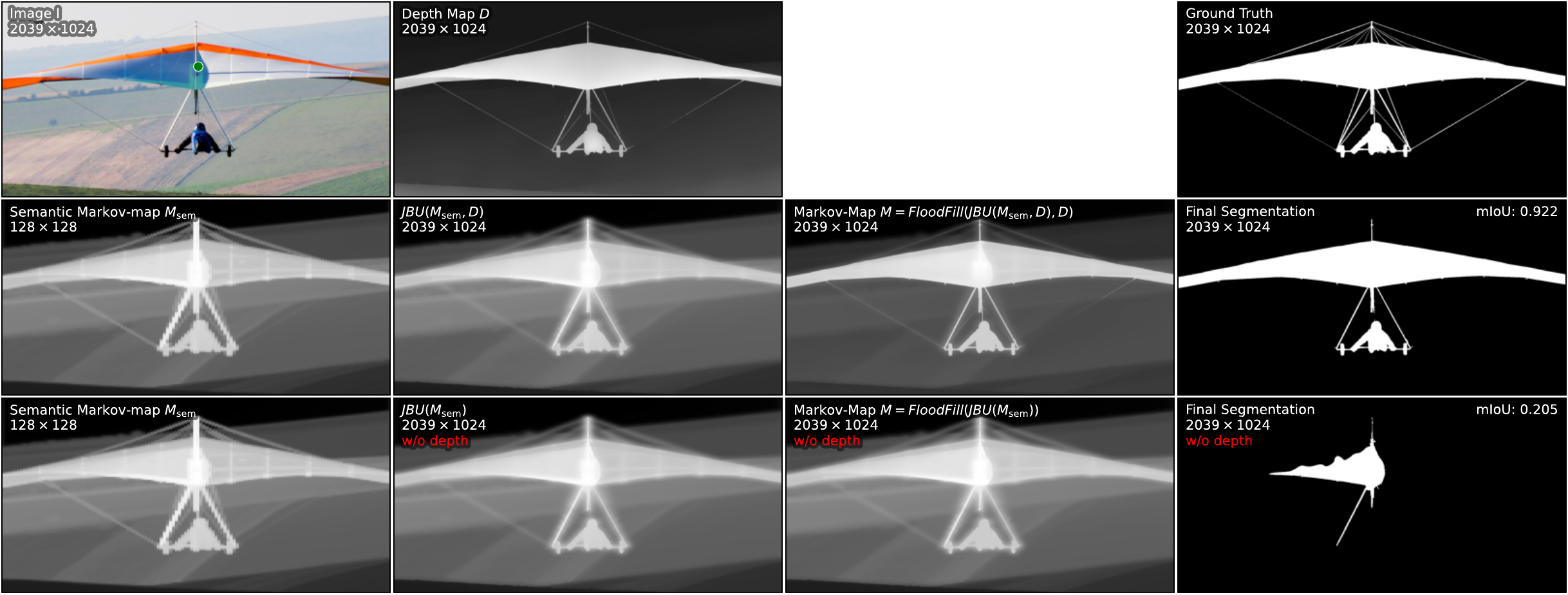}}
  \caption{\textbf{Single Point Depth Guidance Background Separation Example}. The depth map $D$ provides very useful information, if the target instance is clearly in front of the background. We notice a stronger contrast between foreground and background after the flood fill (3rd column) in the depth guided Markov-maps compared to Markov-maps without the depth information.}
  \label{fig:depth_guidance_glider_example}
\end{figure*}

\begin{figure*}
  \centering{\includegraphics[width=\textwidth]{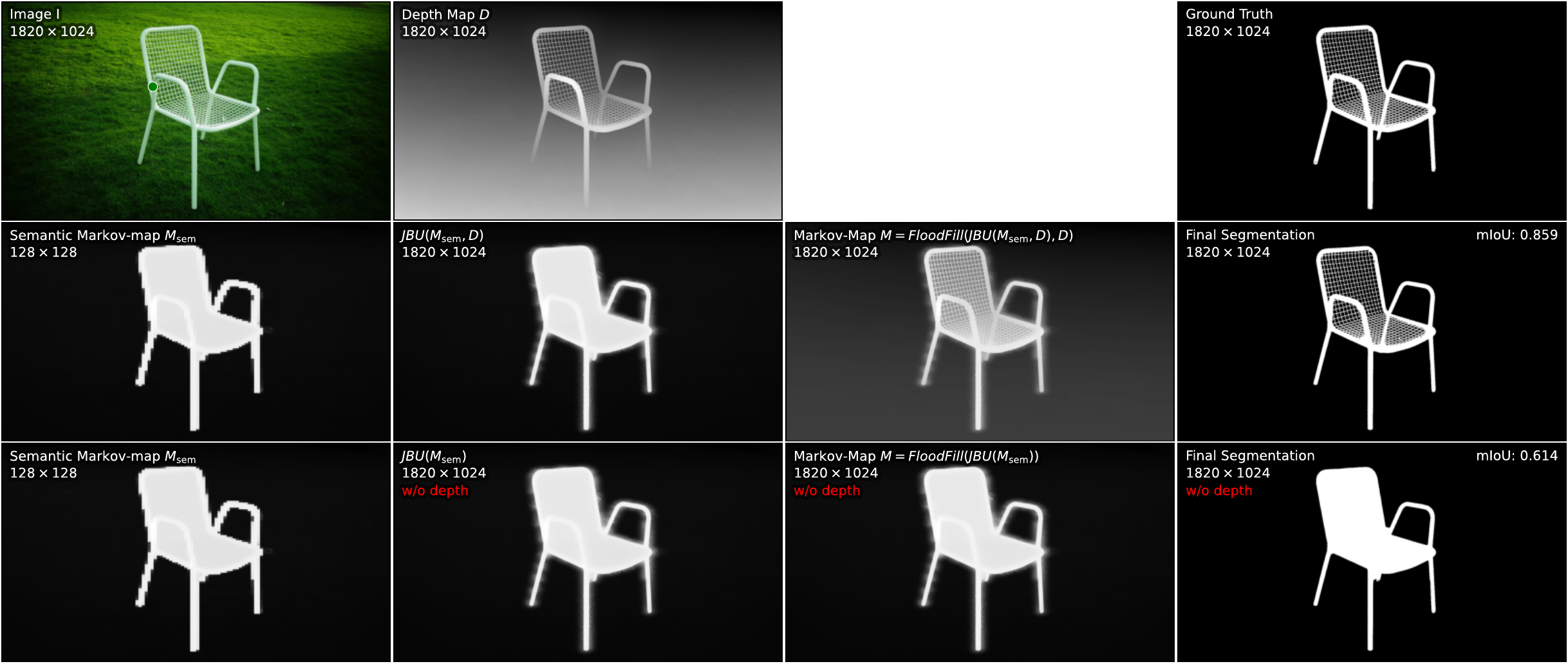}}
  \caption{\textbf{Single Point Depth Guidance Fine Detail Example}. In this example, the Depth map $D$ enables segmenting the net on the chair in the Flood Fill stage (3rd column). Here the version without depth fails, because the semantic Markov-map's $M_\mathrm{sem}$ (1st column) resolution is to low to capture the fine detail and the JBU upscaling (2nd column) can only improve existing edges.}
  \label{fig:depth_guidance_chair_example}
\end{figure*}

\begin{figure*}
  \centering{\includegraphics[width=\textwidth]{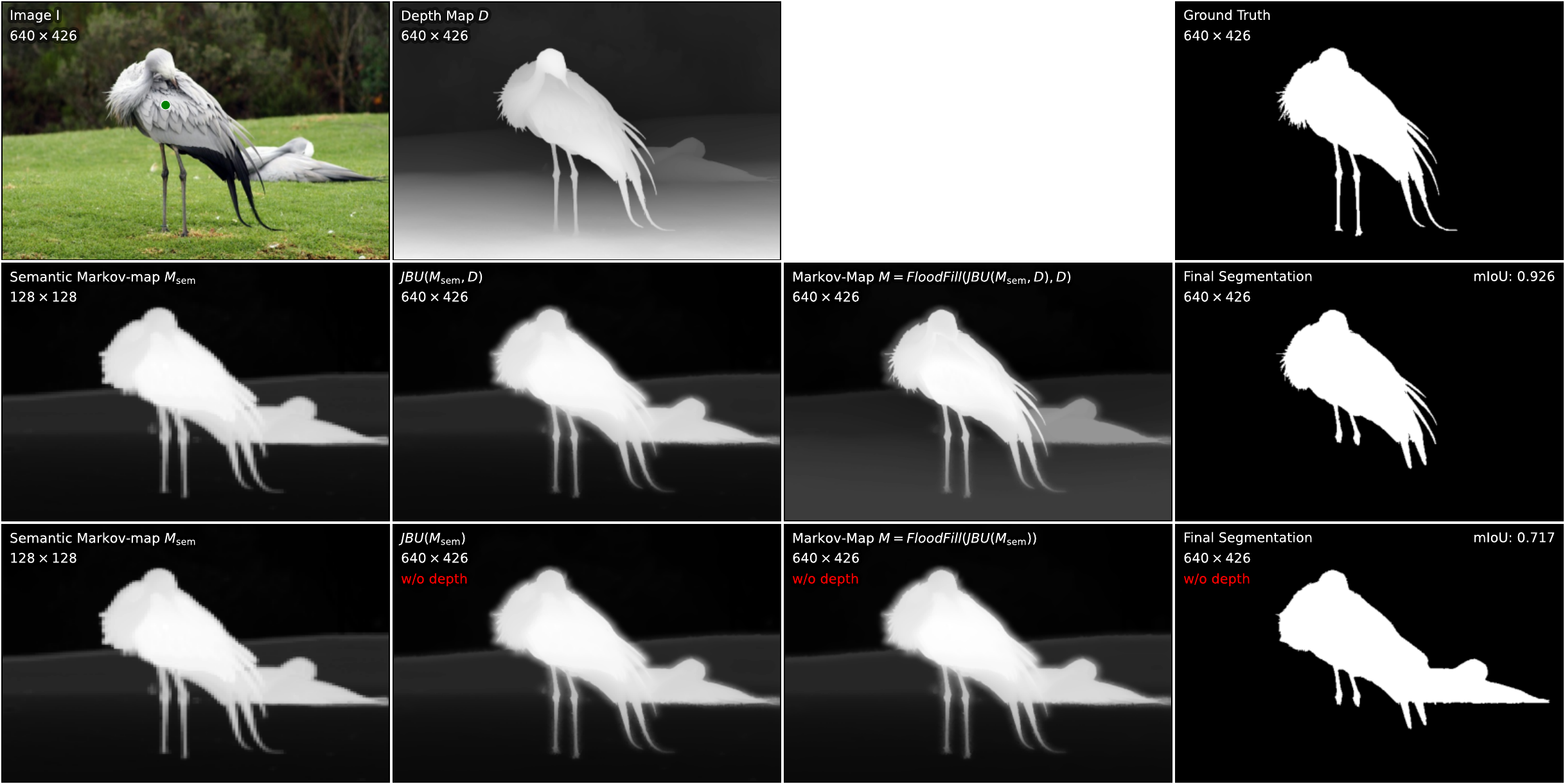}}
  \caption{\textbf{Single Point Depth Guidance Overlapping Instances Example}. A single Markov-map of M2N2 is not able to distinguish between overlapping instances of the same semantic class, as the semantic Markov-maps $M_\mathrm{sem}$ (1st row) only provide semantic information. By using depth, M2N2V2' depth guided Markov-maps are able to separate instances, as the overlapping instances have a different distance to the camera.}
  \label{fig:depth_guidance_example}
\end{figure*}

\begin{figure*}
  \centering{\includegraphics[width=\textwidth]{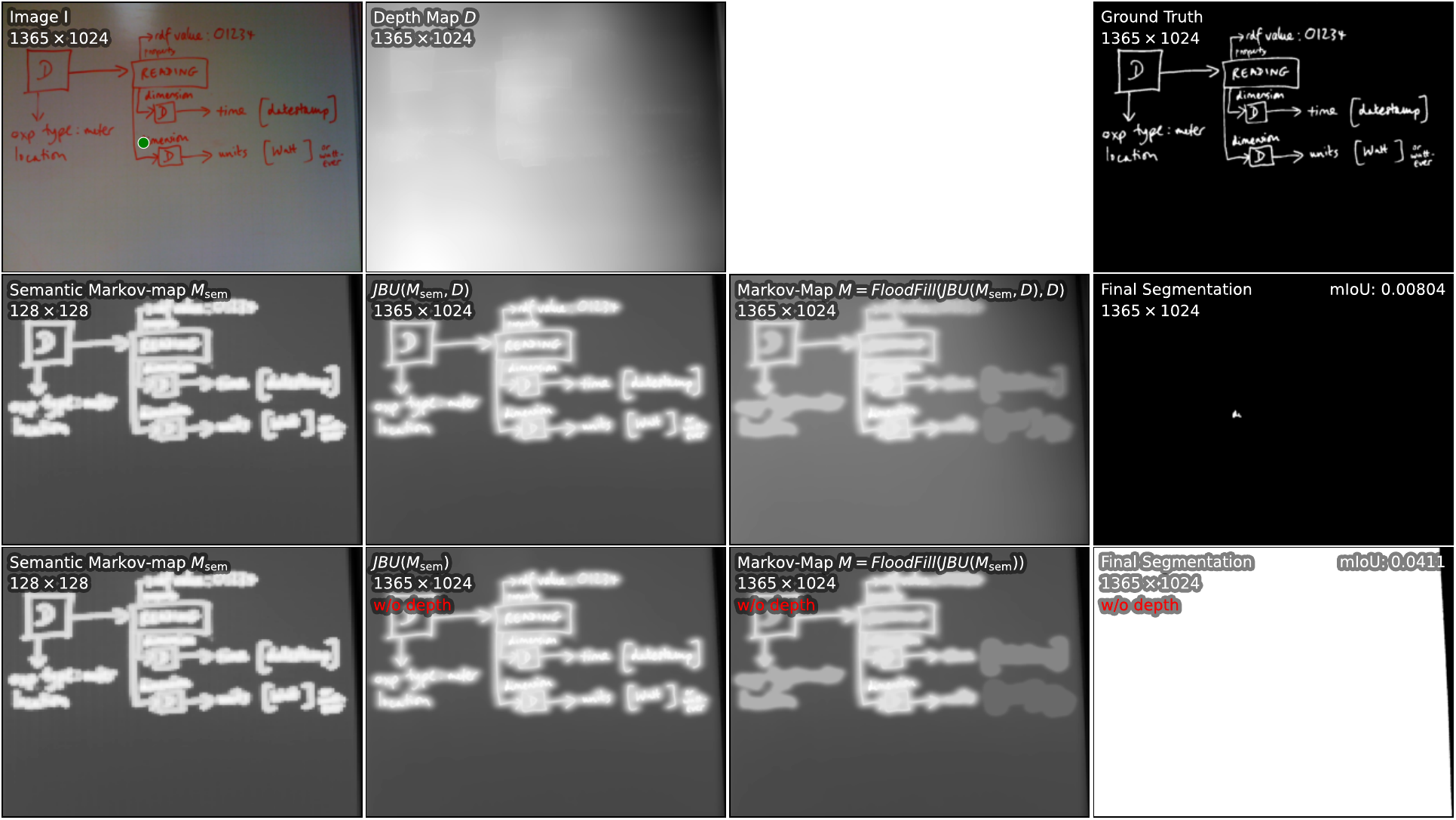}}
  \caption{\textbf{Single Point Depth Guidance Failure Case Example}. In this example, the target segmentation is on a flat whiteboard. M2N2V2 can therefore not take advantage of the depth map, as it does not provide any useful information. We also note that the Flood Fill approach (3rd columns) is suppressing some of the text, as the writing is not fully connected. Furthermore, since there are no clear edges in the Markov-map $M$ (3rd column), The score function fails to find a good scale $\lambda^*$. resulting in either segmenting the entire board or just a single letter.}
  \label{fig:depth_guidance_whiteboard_example}
\end{figure*}

\section{Depth Guidance Examples}\label{sec:depth_guidance_in_detail}
In \cref{fig:depth_guidance_glider_example} we show an example of improving the segmentation by depth guidance, especially in case the object of interest is clearly in foreground. \cref{fig:depth_guidance_chair_example} shows an example of capturing fine details due to depth guidance. In \cref{fig:depth_guidance_example} we observe an example of improved instance segmentation due to depth guidance, where the bird closer to the camera is separated well despite having an overlap with the other bird in the background. Finally, in \cref{fig:depth_guidance_whiteboard_example} we see a failure case where depth guidance does not help to achieve a better segmentation because all objects sharing the same depth.

\begin{figure*}
  \centering{
  \includegraphics[width=\textwidth]{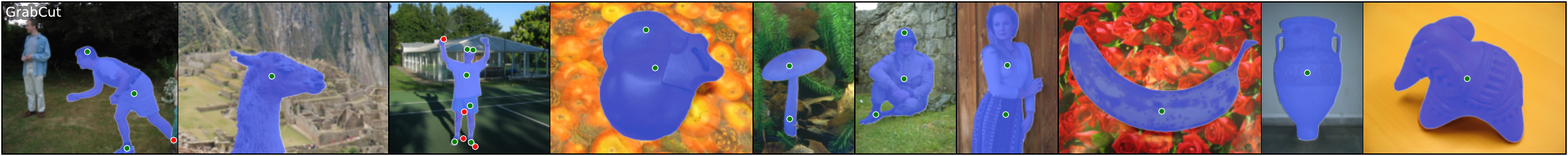}
  \includegraphics[width=\textwidth]{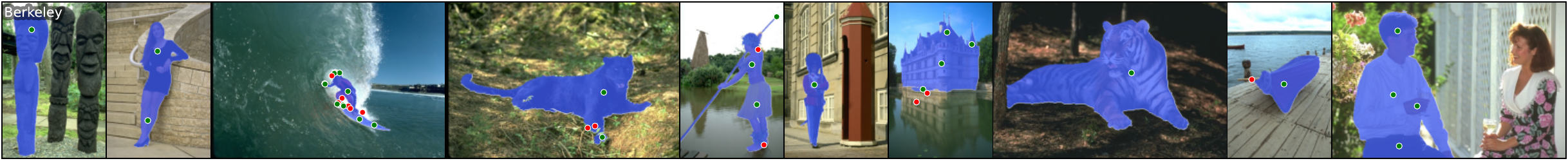}
  \includegraphics[width=\textwidth]{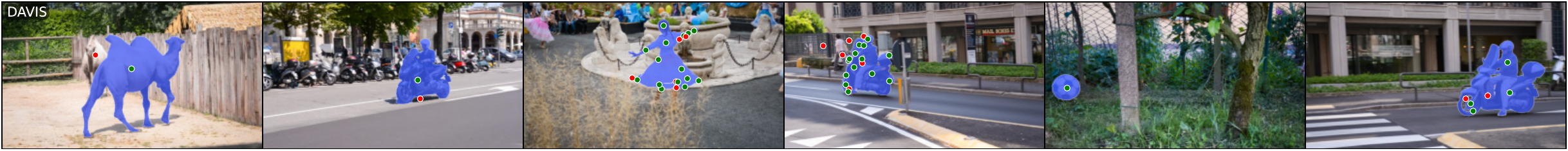}
  \includegraphics[width=\textwidth]{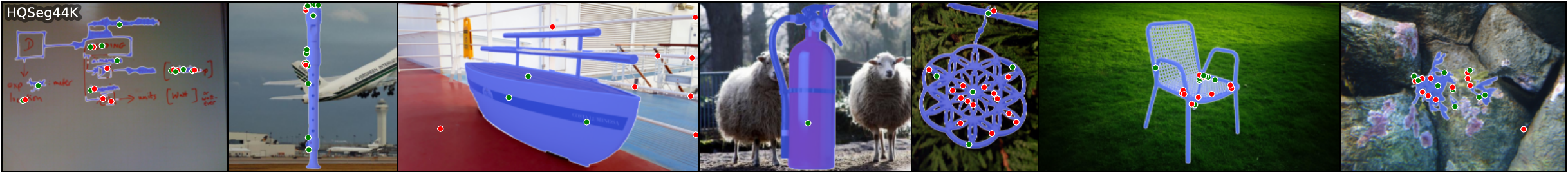}
  \includegraphics[width=\textwidth]{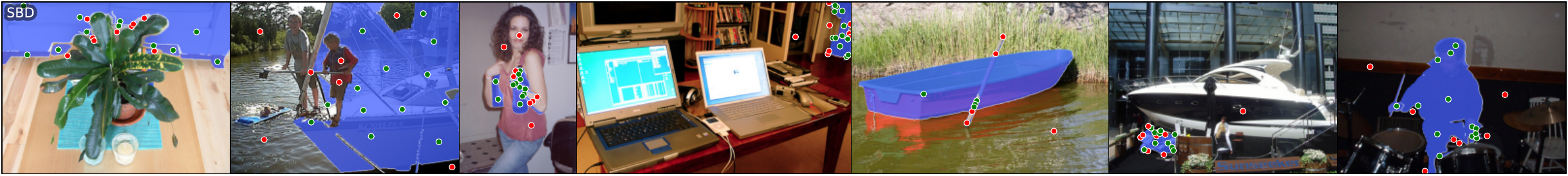}
  \includegraphics[width=\textwidth]{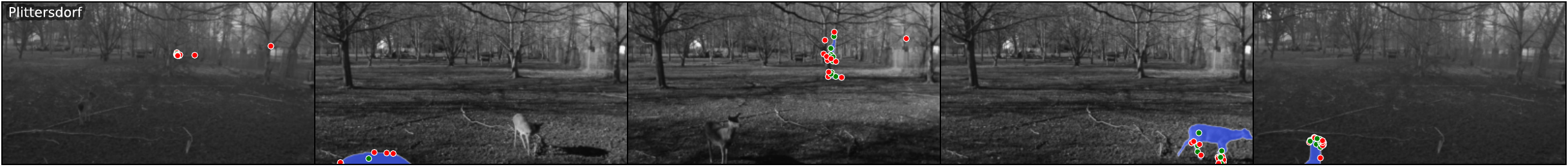}
  \includegraphics[width=\textwidth]{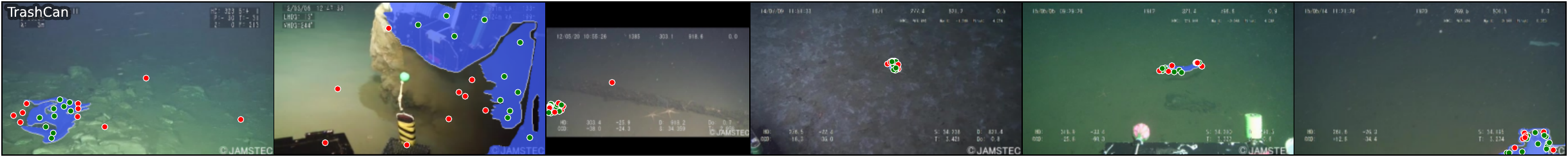}
  \includegraphics[width=\textwidth]{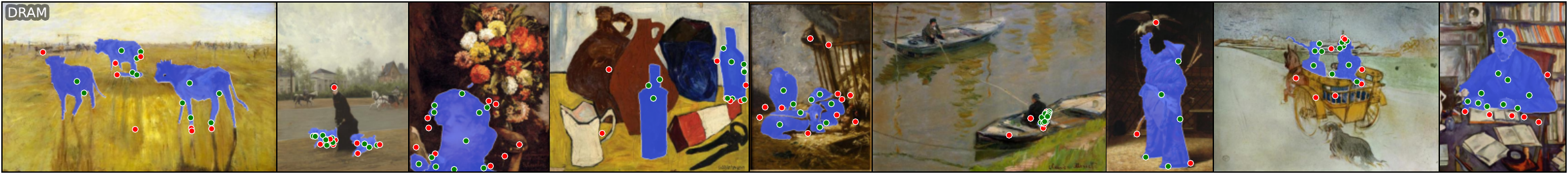}
  \includegraphics[width=\textwidth]{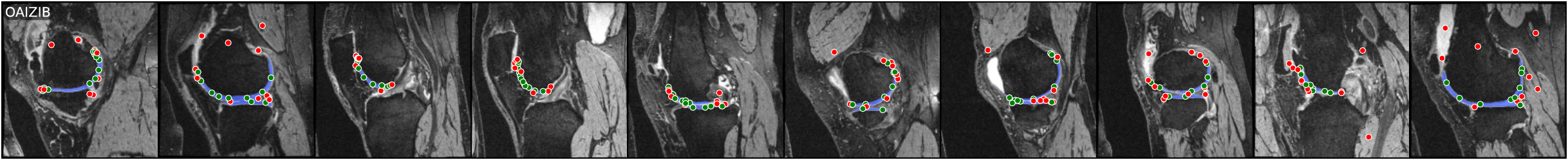}
  \includegraphics[width=\textwidth]{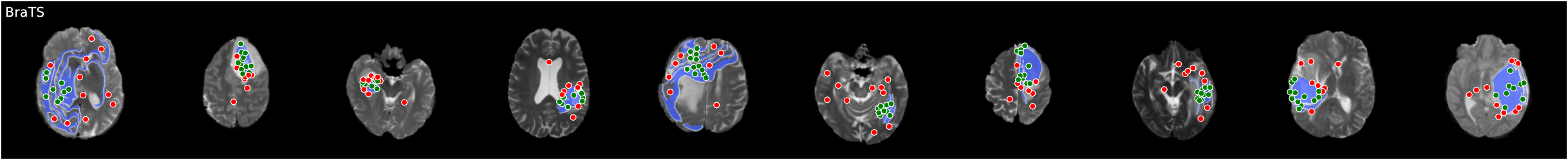}
  }
  \caption{\textbf{Random Predictions of each Dataset}.}
  \label{fig:random_predictions}
\end{figure*}